\documentclass[11pt,a4paper]{article}
\usepackage[hyperref]{acl2020}

\newcommand{\datasetName}{{\sc InfoTabS}\xspace}
\newcommand{\paperTitle}{\datasetName: Inference on Tables as Semi-structured Data}
\newcommand{\unibigrammodelname}{SVM}

\newcommand{\alphaOne}{$\alpha_1$\xspace}
\newcommand{\alphaTwo}{$\alpha_2$\xspace}
\newcommand{\alphaThree}{$\alpha_3$\xspace}

\usepackage{times}
\usepackage{latexsym}

\usepackage{microtype}

\usepackage{graphicx}

\usepackage{subcaption}
\usepackage{url}
\usepackage{amsmath, amsthm, amssymb}

\usepackage{paralist}

\usepackage{enumitem}
\usepackage{xspace}

\usepackage{booktabs} 

\usepackage{wasysym} 
\usepackage{array}

\newcommand{\example}[1]{\emph{#1}}

\usepackage{lingmacros}

\aclfinalcopy 
\def\aclpaperid{2363} 

\title{\paperTitle}

\author{Vivek Gupta, Maitrey Mehta, Pegah Nokhiz, Vivek Srikumar \\
  School of Computing, University of Utah \\
  \texttt{\{vgupta,maitrey,pnokhiz,svivek\}@cs.utah.edu 
  }\\}

\date{}

\newcommand{\skippedDetails}[1]{}

\begin{document}
\maketitle

\begin{abstract}
  In this paper, we observe that semi-structured tabulated text is ubiquitous;
  understanding them requires not only comprehending the meaning of text
  fragments, but also implicit relationships between them. We argue that such
  data can prove as a testing ground for understanding how we reason about
  information.
  To study this, we introduce a new dataset called \datasetName, comprising of
  human-written textual hypotheses based on premises that are tables extracted
  from Wikipedia info-boxes. Our analysis shows that the semi-structured,
  multi-domain and heterogeneous nature of the premises admits complex,
  multi-faceted reasoning.
  Experiments reveal that, while human annotators agree on the relationships
  between a table-hypothesis pair, several standard modeling strategies are
  unsuccessful at the task, suggesting that reasoning about tables can pose a
  difficult modeling challenge. 
 
\end{abstract}

\section{Introduction}
\label{sec:intro}

Recent progress in text understanding has been driven by sophisticated neural
networks based on contextual embeddings---e.g., BERT~\cite{devlin2019bert}, and
its descendants---trained on massive datasets, such as
SNLI~\cite{snli:emnlp2015}, MultiNLI~\cite{N18-1101}, and
SQuAD~\cite{rajpurkar-etal-2016-squad}. Several such models outperform human
baselines on these tasks on the benchmark suites such as GLUE~\cite{wang2018glue}.
Reasoning about text requires a broad array of skills---making lexical
inferences, interpreting the nuances of time and locations, and accounting for
world knowledge and common sense. \emph{Have we achieved human-parity across
  such a diverse collection of reasoning skills?}

\begin{figure}[t]
  \centering
  {
  \footnotesize
  \begin{center}
    \begin{tabular}{>{\raggedright}p{0.3\linewidth}p{0.45\linewidth}}
      \multicolumn{2}{c}{\bf Dressage}                                                    \\
      \toprule
      {\bf Highest governing body} & International Federation for Equestrian Sports (FEI) \\
      \midrule
      \multicolumn{2}{c}{\emph{Characteristics}}                                          \\
      {\bf Contact}                & No                                                   \\
      {\bf Team members}           & Individual and team at international levels          \\
      {\bf Mixed gender}           & Yes                                                  \\
      {\bf Equipment}              & Horse, horse tack                                   \\
      {\bf Venue}                  & Arena, indoor or outdoor                             \\
      \midrule
      \multicolumn{2}{c}{\emph{Presence}}                                                 \\
      {\bf Country or region}      & Worldwide                                            \\
      {\bf Olympic}                & 1912                                                 \\
      {\bf Paralympic}             & 1996                                                 \\
      \bottomrule
    \end{tabular}
  \end{center}
}
  {\footnotesize
    \begin{enumerate}[nosep]
    \item[H1:] Dressage was introduced in the Olympic games in 1912.
    \item[H2:] Both men and women compete in the equestrian sport of Dressage.
    \item[H3:] A dressage athlete can participate in both individual and team events.
    \item[H4:] FEI governs dressage only in the U.S. 
    \end{enumerate}
  }
    \caption{A semi-structured premise (the table). Two hypotheses (H1, H2)
    are entailed by it, H3 is neither entailed nor contradictory, and H4 is a contradiction. }
  \label{fig:example}
\end{figure}

In this paper, we study this question by proposing an extension of the natural language
inference (NLI) task~\cite[][and others]{dagan2005pascal}. In NLI, which asks whether a premise
entails, contradicts or is unrelated to a hypothesis, the premise and the
hypothesis are one or more sentences. Understanding the
premise requires understanding its linguistic structure and reasoning about
it. We seek to separate these two components. Our work stems from the
observation that we can make valid inferences about \emph{implicit} information
conveyed by the mere juxtaposition of snippets of text, as shown in the table
describing \example{Dressage} in Figure~\ref{fig:example}.

We introduce the \datasetName dataset to study and model inference with such
semi-structured data. Premises in our dataset consist of info-boxes that convey
information implicitly, and thus require complex reasoning to ascertain the
validity of hypotheses. For example, determining that the hypothesis H2 in
Figure~\ref{fig:example} entails the premise table requires looking at multiple
rows of the table, understanding the meaning of the row labeled \example{Mixed gender},
and also that \example{Dressage} is a sport.

\datasetName consists of 23,738 premise-hypothesis pairs, where all
premises are info-boxes, and the hypotheses are short sentences. As in the NLI
task, the objective is to ascertain whether the premise entails, contradicts or
is unrelated to the hypothesis. The dataset
has 2,540 unique info-boxes drawn from Wikipedia articles across various categories,
and all the hypotheses are written by Amazon's Mechanical Turk workers. Our
analysis of the data shows that ascertaining the label typically requires the
composing of multiple types of inferences across multiple rows from the
tables in the context of world knowledge. Separate verification experiments on
subsamples of the data also confirm the high quality of the dataset.

We envision our dataset as a challenging testbed for studying how models can
reason about semi-structured information. To control for the possibility of
models memorizing superficial similarities in the data to achieve high
performance, in addition to the standard train/dev/test split, our dataset
includes two additional test sets that are constructed by systematically
changing the surface forms of the hypothesis and the domains of the tables. We
report the results of several families of approaches representing word overlap
based models, models that exploit the structural aspect of the premise, and also
derivatives of state-of-the-art NLI systems. Our experiments reveal that all
these approaches underperform across the three test sets.

In summary, our contributions are:
\begin{enumerate}[nosep]
\setlength\itemsep{0.25em}
\item We propose a new English natural language inference dataset, \datasetName, to study the problem of reasoning about semi-structured data.
  
\item To differentiate models' ability to reason about the premises from their
  memorization of spurious patterns, we created three challenge test sets
  with controlled differences that employ similar reasoning as the training set.
  
\item We show that several existing approaches for NLI underperform on our
  dataset, suggesting the need for new modeling strategies.
  
\end{enumerate}

\noindent The dataset, along with associated scripts, are available at \url{https://infotabs.github.io/}.

\section{The Case for Reasoning about Semi-structured Data}
\label{sec:reasoning-about-tables}

We often encounter textual information that is neither unstructured (i.e., raw
text) nor strictly structured (e.g., databases). Such data, where a structured
scaffolding is populated with free-form text, can range from the highly verbose
(e.g., web pages) to the highly terse (e.g. fact sheets, information tables,
technical specifications, material safety sheets). Unlike databases, such
semi-structured data can be heterogeneous in nature, and not characterized by
pre-defined schemas. Moreover, we may not always have accompanying explanatory
text that provides context. Yet, we routinely make inferences about such heterogeneous, incomplete
information and fill in gaps in the available information using our expectations
about relationships between the elements in the data.

Understanding semi-structured information requires a broad spectrum of reasoning
capabilities. We need to understand information in an ad hoc layout constructed
with elements (cells in a table) that are text snippets, form fields or are
themselves sub-structured (e.g., with a list of elements). Querying such data
can require various kinds of inferences. At the level of individual cells, these
include simple lookup (e.g., knowing that \example{dressage takes place in an
  arena}), to lexical inferences (e.g., understanding that \example{Mixed
  Gender} means both men and women compete), to understanding types of text
in the cells (e.g., knowing that the number 1912 is a year). Moreover, we may
also need to aggregate information across multiple rows (e.g., knowing that
\emph{dressage is a non-contact sport that both men and women compete in}), or
perform complex reasoning that combines temporal information with world
knowledge.

We argue that a true test of reasoning should evaluate the ability to handle
such semi-structured information. To this end, we define a new task modeled
along the lines of NLI, but with tabular premises and textual hypotheses, and
introduce a new dataset \datasetName for this task.


\section{The Need for Multi-Faceted Evaluation}
\label{sec:artifact}

Before describing the new dataset, we will characterize our approach for a
successful evaluation of automated reasoning.

Recent work has shown that many datasets for NLI contain annotation biases or artifacts~\cite[e.g.][]{poliak-etal-2018-hypothesis}. In other words, large models trained on such datasets are prone to learning spurious patterns---they can predict correct labels even with incomplete or noisy inputs. For instance, \example{not} and \example{no} in a hypothesis are correlated with contradictions~\cite{niven-kao-2019-probing}.  Indeed, classifiers trained on the hypotheses only (ignoring the premises completely) report high accuracy; they exhibit \emph{hypothesis bias}, and achieving a high predictive performance does not need models to discover relationships between the premise and the hypothesis. Other artifacts are also possible. For example, annotators who generate text may
use systematic patterns that ``leak'' information about the label to a
model. Or, perhaps models can learn correlations that mimic reasoning, but only
for one domain.  With millions of parameters, modern neural networks are prone
to overfitting to such imperceptible patterns in the data.

From this perspective, if we seek to measure a model's capability to understand
and reason about inputs, we cannot rely on a single fixed test set to rank
models. Instead, we need multiple test sets (of similar sizes) that have
controlled differences from each other to understand how models handle changes
along those dimensions. While all the test sets address the same task, they may
not all be superficially similar to the training data. 

With this objective,  we build three test sets, named \alphaOne,
\alphaTwo and \alphaThree. Here, we briefly introduce them;
\S\ref{sec:construction} goes into specifics.
Our first test set (\alphaOne) has a similar distribution as the training data
in terms of lexical makeup of the hypotheses and the premise domains.

The second,  \emph{adversarial test set (\alphaTwo)}, consists of examples that are also similar in distribution to the training set, but the hypothesis labels are changed by expert annotators changing as few words in the sentence as possible. For instance, if \example{Album $X$ was released in the $21^{st}$ century} is an entailment, the sentence  \example{Album $X$ was released before the $21^{st}$ century}  is a contradiction, with only one change. Models that merely learn superficial textual artifacts will get confused by the new sentences. For \alphaTwo, we rewrite entailments as contradictions and vice versa, while the neutrals are left unaltered. 

Our third test set is the \emph{cross-domain (\alphaThree)} set, which uses
premises from domains that are not in the training split, but generally,
necessitate similar types of reasoning to arrive at the entailment
decision. 
Models that overfit domain-specific artifacts will underperform on \alphaThree.

Note that, in this work, we describe and introduce three different test sets,
but we expect that future work can identify additional dimensions along which
models overfit their training data and construct the corresponding test sets.


\section{The \datasetName Dataset}
\label{sec:construction}
In this section, we will see the details of the construction of \datasetName. We
adapted the general workflow of previous crowd sourcing approaches for creating
NLI tasks~\cite[e.g.,][]{snli:emnlp2015} that use Amazon's Mechanical Turk.\footnote{Appendix \ref{sec:data_examples} has more examples of tables with hypotheses.}

\paragraph{Sources of Tables}
Our dataset is based on $2,540$ unique info-boxes from Wikipedia articles across multiple categories (listed in Appendix~\ref{sec:generation_stats_appendix}). 
We did not include tables that have fewer than 3 rows, or have non-English cells (e.g., Latin names of plants) and  technical information that may require expertise to understand (e.g., astronomical details about exoplanets). We also removed non-textual information from the table, such as images. Finally, we simplified large tables into smaller ones by splitting them at sub-headings. Our tables are isomorphic to key-value pairs, e.g., in Figure \ref{fig:example}, the bold entries are the keys, and the corresponding entries in the same row are their respective values.

\paragraph{Sentence generation}
Annotators were presented with a tabular premise and instructed to write three
self-contained grammatical sentences based on the tables: one of which is
true given the table, one which is false, and one which may or may not be
true. The turker instructions included illustrative examples using a table
and also general principles to bear in mind, such as avoiding information that is not
widely known, and avoiding using information that is not in the table (including
names of people or places). The turkers were encouraged not to restate information in the table, or make trivial changes such as the addition
of words like \example{not} or changing numerical values. We refer the reader to the project website for a snapshot of the interface used for turking, which includes the details of instructions. 

We restricted the turkers to be from English-speaking countries with at least a
Master's qualification. We priced each HIT (consisting of one table) at
$50\cent$. Following the initial turking phase, we removed grammatically bad
sentences and rewarded workers whose sentences involved multiple rows in the
table with a $10\%$ bonus. Appendix~\ref{sec:appendix_worker_stat} gives
additional statistics about the turkers.

\paragraph{Data partitions}
We annotated $2,340$ unique tables with nine sentences per table (i.e., three turkers
per table).\footnote{For tables with ungrammatical sentences, we repeated the HIT.\@ As a result, a few tables in the final data release have more than $9$ hypotheses.} We partitioned these tables into training, development (Dev), \alphaOne and \alphaTwo test sets. To prevent an outsize impact of influential turkers in a split, we ensured that the annotator distributions in the Dev and test splits are similar to that of the training split.

We created the \alphaTwo test set from hypotheses similar to those in \alphaOne, but from a separate set of tables, and perturbing them as described in \S\ref{sec:artifact}. On an average, $\sim 2.2$ words were changed per sentence to create \alphaTwo, with no more than $2$ words changing in $72\%$ of the hypotheses. The provenance of \alphaTwo ensures that the kinds of reasoning needed for \alphaTwo are similar to those in \alphaOne and the development set.
For the \alphaThree test set, we annotated 200 
additional tables belonging to domains not seen in the training set (e.g., diseases, festivals). As we will see in \S\ref{sec:main-reasoning}, hypotheses in these categories involve a set of similar types of reasonings as \alphaOne, but with different distributions.

\begin{table}
  \centering
  \begin{tabular}{lcc}
    \toprule
    Data split       & \# tables & \# pairs \\
    \midrule
    Train            & 1740        & 16538       \\
    Dev              & 200        & 1800       \\
    \alphaOne test   & 200        & 1800       \\
    \alphaTwo test   & 200        & 1800       \\
    \alphaThree test & 200        & 1800       \\  
    \bottomrule
  \end{tabular}
  \caption{Number of tables and premise-hypothesis pairs for each data split}
  \label{tab:data-splits-stats}
\end{table}

In total, we collected $23,738$ sentences split almost equally among entailments, contradictions, and neutrals. Table~\ref{tab:data-splits-stats} shows the number of tables and premise-hypothesis pairs in each split.
In all the splits, the average length of the hypotheses is similar. We refer the reader to Appendix~\ref{sec:generation_stats_appendix} for additional statistics about the data.

\paragraph{Validating Hypothesis Quality}
We validated the quality of the data using Mechanical Turk. For each premise-hypothesis in the development and the test sets, we asked turkers to predict whether the hypothesis is entailed or contradicted by, or is unrelated to the premise table. We priced this task at $36\cent$ for nine labels. 

The inter-annotator agreement statistics are shown in Table~\ref{tab:verification_statistics}, with detailed statistics in Appendix~\ref{sec:appendix_verification}. On all splits, we observed significant inter-annotator agreement scores with Cohen's Kappa scores~\cite{artstein2008inter} between $0.75$ and $0.80$. In addition, we see a majority agreement (at least $3$ out of $5$ annotators agree) of range between 93$\%$ and 97$\%$. Furthermore, the human accuracy agreement between the majority and gold label (i.e., the label intended by the writer of the hypothesis), for all splits is in range 80$\%$ to 84$\%$, as expected given the difficulty of the task.  

\begin{table}
    \centering
    \begin{tabular}{lccc}
    \toprule
    \bf Dataset   & \bf Cohen's & \bf Human    & \bf Majority  \\ 
                  & \bf Kappa & \bf Accuracy & \bf Agreement \\
      \midrule      
      Dev         & 0.78      & 79.78        & 93.52         \\ 
      \alphaOne   & 0.80      & 84.04        & 97.48         \\ 
      \alphaTwo   & 0.80      & 83.88        & 96.77         \\ 
      \alphaThree & 0.74      & 79.33        & 95.58         \\
      \bottomrule
    \end{tabular}
    \caption{Inter-annotator agreement statistics}        
    \label{tab:verification_statistics}
\end{table}


\section{Reasoning Analysis} 
\label{sec:main-reasoning}

To study the nature of reasoning that is involved in deciding the relationship
between a table and a hypothesis, we adapted the set of reasoning categories
from GLUE~\cite{wang2018glue} to table premises.
For brevity, here we will describe the categories that are not in GLUE and defined in this work for table premises. Appendix \ref{sec:reasoning} gives the full list with definitions and examples.
\emph{Simple look up} refers to cases where there is no reasoning and the hypothesis is formed by literally restating what is in the table as a sentence; \emph{multi-row reasoning} requires multiple rows to make an inference; and \emph{subjective/out-of-table} inferences involve value judgments about a proposition or reference to information out of the table that is neither well known or common sense. 

\begin{figure*}
        \centering
        \begin{subfigure}[b]{0.46\textwidth}
            \centering
            \includegraphics[width=\textwidth]{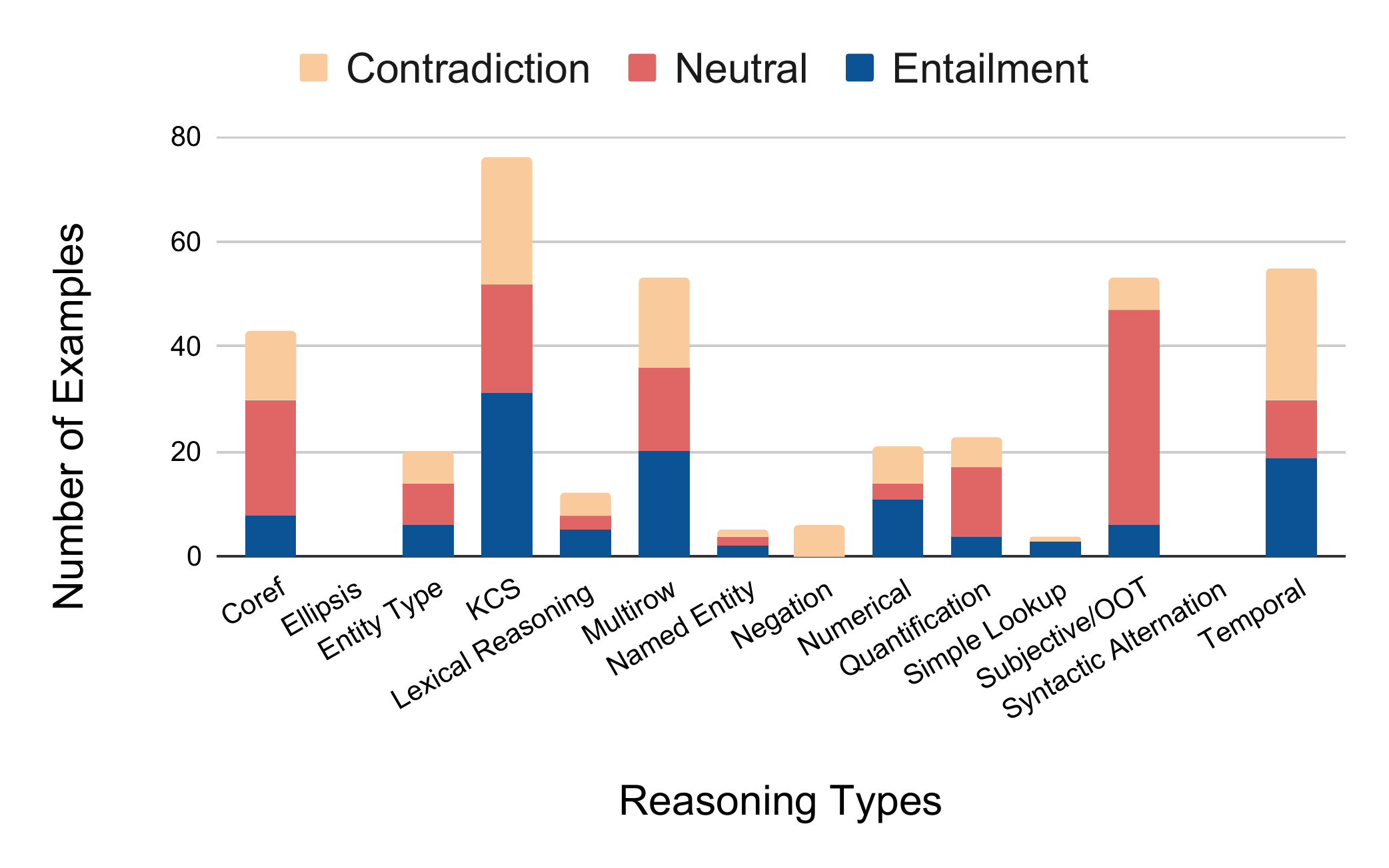}
            \caption{Number of examples per reasoning type in the Dev set}
            \label{fig:reasoning_dev}
        \end{subfigure}
        \hfill
        \begin{subfigure}[b]{0.46\textwidth}  
            \centering 
            \includegraphics[width=\textwidth]{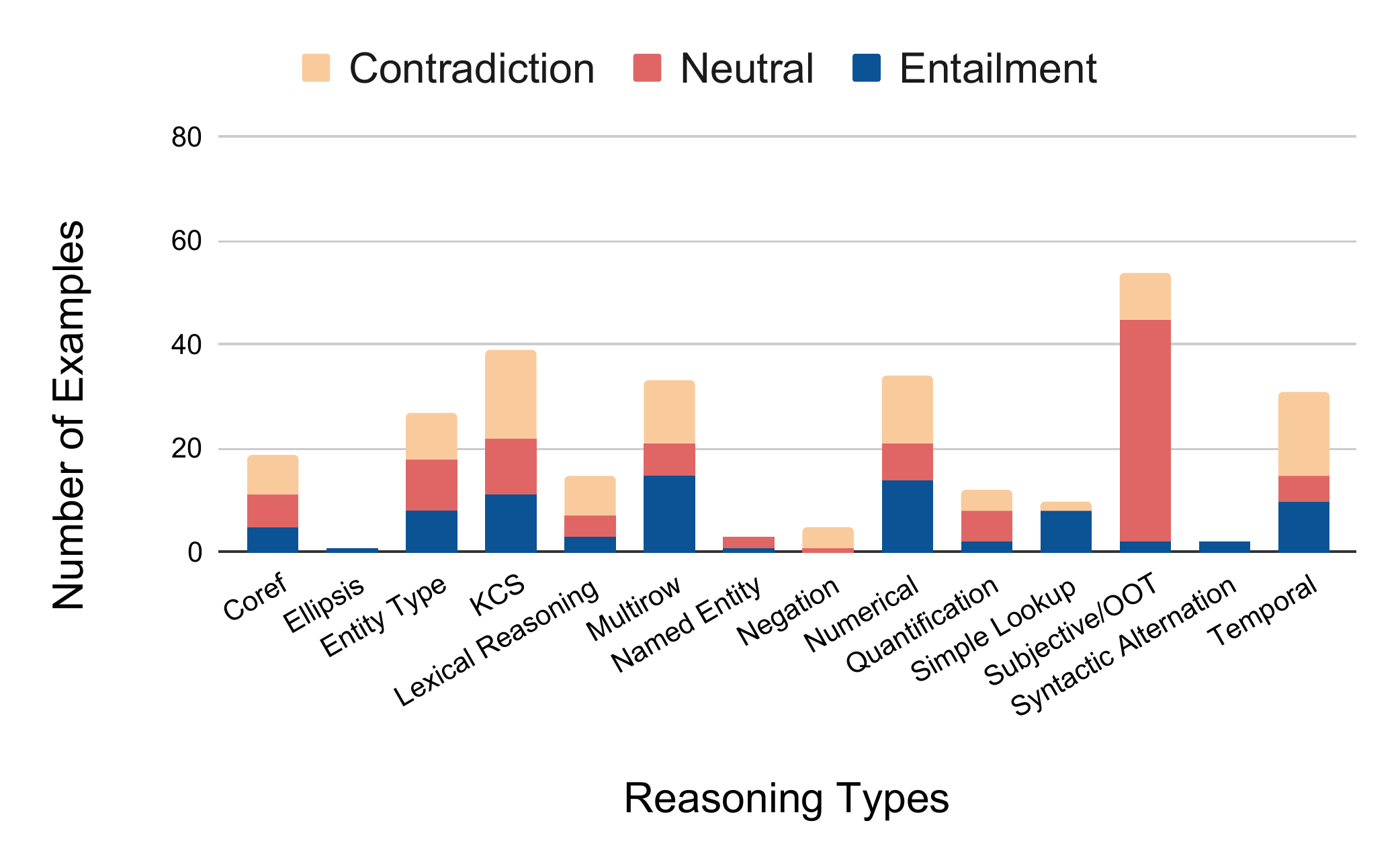}
            \caption{Number of examples per reasoning type in the \alphaThree~ set}
            \label{fig:reasoning_alphaThree}
        \end{subfigure}
        \vskip\baselineskip
        \begin{subfigure}[b]{0.46\textwidth}   
            \centering 
            \includegraphics[width=\textwidth]{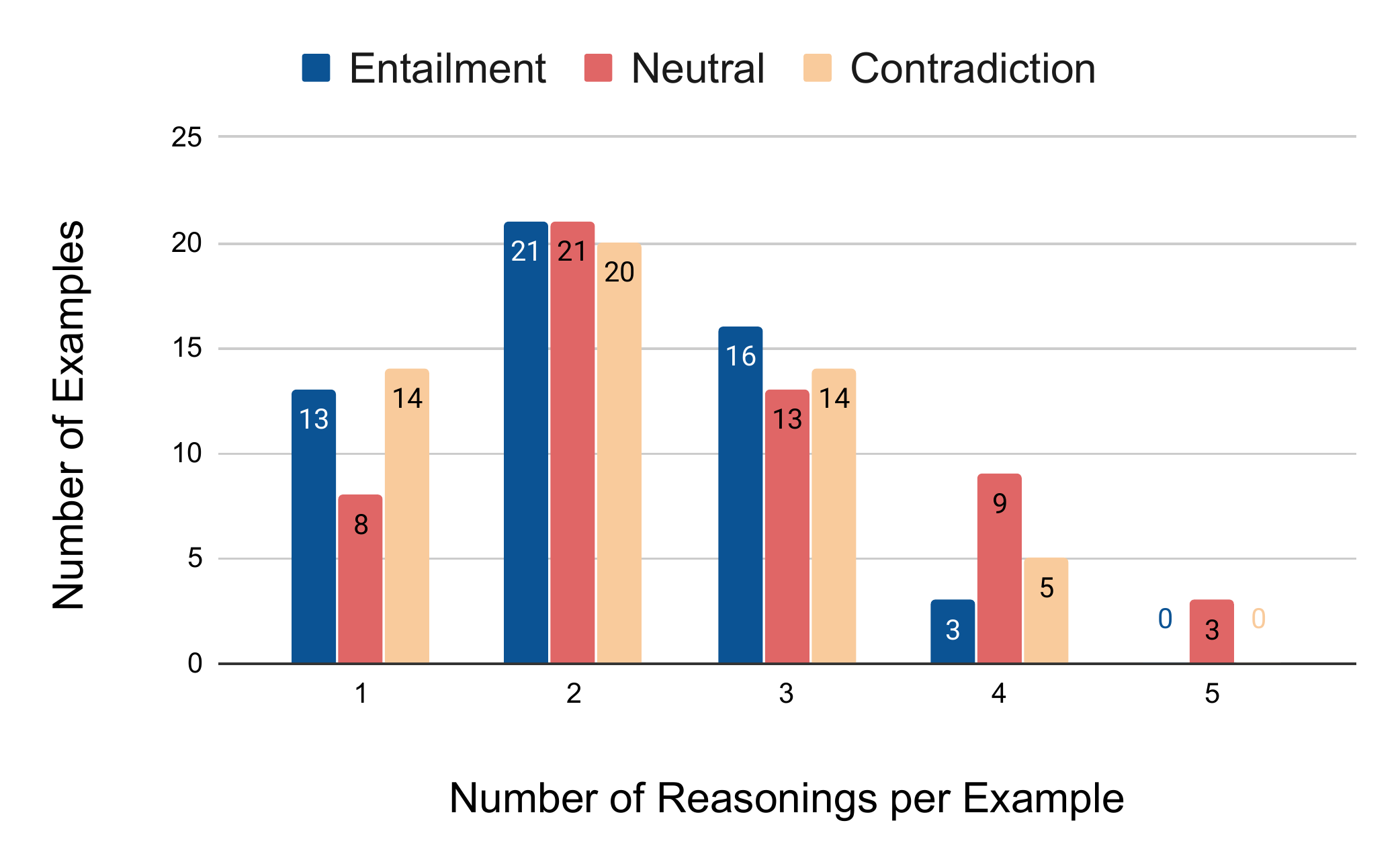}
            \caption{Number of reasonings per example in the Dev set}
            \label{fig:reasoning_dev_no}
        \end{subfigure}
        \quad
        \begin{subfigure}[b]{0.46\textwidth}   
            \centering 
            \includegraphics[width=\textwidth]{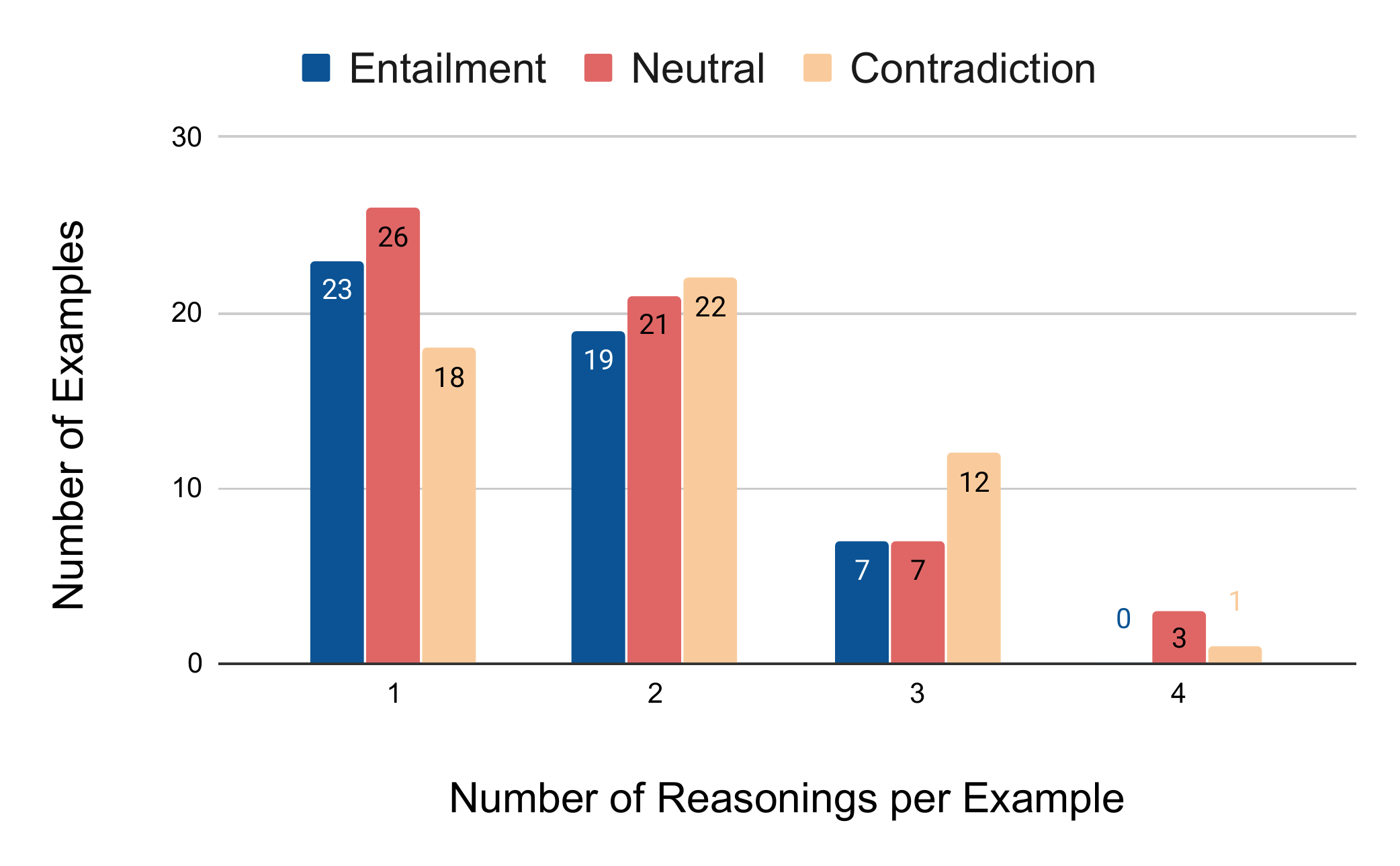}
        \caption{Number of reasonings per example in the \alphaThree~ set}
        \label{fig:reasoning_alphaThree_no}
        \end{subfigure}
        \caption{Distribution of the various kinds of reasoning in the Dev and \alphaThree sets. The labels OOT and KCS are short for out-of-table and Knowledge \& Common Sense, respectively.}
        \label{fig:reasoning statistics}
    \end{figure*}

All definitions and their boundaries were verified via several rounds of discussions. Following this, three graduate students independently annotated 160 pairs from the Dev and \alphaThree test sets each, and edge cases were adjudicated to arrive at consensus labels.
Figures \ref{fig:reasoning_dev} and \ref{fig:reasoning_alphaThree} summarizes these annotation efforts. We see that we have a multi-faceted complex range of reasoning types across both sets. Importantly, we observe only a small number of simple lookups, simple negations for contradictions, and mere syntactic alternations that can be resolved without complex reasoning. Many instances call for looking up multiple rows, and involve temporal and numerical reasoning. 
Indeed,  as Figures \ref{fig:reasoning_dev_no} and \ref{fig:reasoning_alphaThree_no} show, a large number of examples need at least two distinct kinds of reasoning;
on an average, sentences in the Dev and \alphaThree sets needed 2.32 and 1.79 different kinds of reasoning, respectively.

We observe that semi-structured premises forced annotators to call upon world knowledge and common sense (KCS); $48.75\%$ instances in the Dev set require KCS. (In comparison, in the MultiNLI data, KCS is needed in $25.72\%$ of examples.) We conjecture that this is because information about the entities and their types is not explicitly stated in tables, and have to be inferred.  To do so, our annotators relied on their knowledge about the world including information about weather, seasons, and widely known social and cultural norms and facts. An example of such common sense is the hypothesis that \example{``$X$ was born in summer''} for a person whose date of birth is in May in New York. We expect that the \datasetName data can serve as a basis for studying common sense reasoning alongside other recent work such as that of~\citet{talmor-etal-2019-commonsenseqa}, 

Neutral hypotheses are more inclined to being subjective/out-of-table because almost anything subjective or not mentioned in the table is a neutral statement.  Despite this, we found that in all evaluations in Appendix \ref{sec:f1score} (except those involving the adversarial \alphaTwo test set), our models found neutrals almost as hard as the other two labels, with only an $\approx 3\%$ gap between the F-scores of the neutral label and the next best label.



The distribution of train, dev, \alphaOne~and \alphaTwo~are similar because the premises are taken from the same categories. However, tables for \alphaThree~are from different domains, hence not of the same distribution as the previous splits. This difference is also reflected in Figures \ref{fig:reasoning_dev} and \ref{fig:reasoning_alphaThree}, as we see a different distribution of reasonings for each test set. This is expected; for instance, we cannot expect temporal reasoning from tables in a domain that does not contain temporal quantities. 

\section{Experiments and Results}
\label{sec:experiments}
The goal of our experiments is to study how well different modeling approaches
address the \datasetName data, and also to understand the impact
of various artifacts on them. First, we will consider different approaches for representing tables in ways that are amenable to modern neural models.

\subsection{Representing Tables}
\label{sec:representations-tables-words}

A key aspect of the \datasetName task that does not apply to the standard
NLI task concerns how premise tables are represented. As baselines for
future work, let us consider several different approaches.

\begin{enumerate}[nosep]
\item \textbf{Premise as Paragraph (Para)}: We convert the premise table into paragraphs using fixed template applied to each row. For a table titled \texttt{t}, a row with key \texttt{k} and value \texttt{v} is written as the sentence \example{The \texttt{k} of \texttt{t} are \texttt{v}.} For example, for the table in Figure~\ref{fig:example}, the row with key \example{Equipment} gets mapped to the sentence \example{The equipment of Dressage are horse, horse tack.}
We have a small number of exceptions: e.g., if the key is \textit{born} or \textit{died}, we use the following template: \example{\texttt{t} was \texttt{k} on \texttt{v}}. 
  
The sentences from all the rows in the table are concatenated to form the premise paragraph. While this approach does not result in grammatical sentences, it fits the interface for standard sentence encoders.

\item \textbf{Premise as Sentence (Sent):} Since hypotheses are typically short, they may be derived from a small subset of rows. Based on this intuition, we use the word mover distance~\cite{kusner2015word} to select the closest and the three closest sentences to the hypothesis from the paragraph representation (denoted by WMD-1 and WMD-3, respectively).

\item \textbf{Premise as Structure 1 (TabFact)}: Following~\citet{2019TabFactA}, we represent tables by a sequence of \texttt{key} \texttt{:} \texttt{value} tokens. Rows are separated by a semi-colon and multiple values for the same key are separated by a comma. 

\item \textbf{Premise as Structure 2 (TabAttn)}: To study an attention based approach, such as that of~\citet{parikh-etal-2016-decomposable}, we convert keys and values into a contextually enriched vectors by first converting them into sentences using the Para approach above, and applying a contextual encoder to each sentence. From the token embeddings, we obtain the embeddings corresponding of the keys and values by mean pooling over only those tokens. 

\end{enumerate}

\subsection{Modeling Table Inferences}
\label{sec:models}
Based on the various representations of tables described above, we developed a collection of models for the table inference problem, all based on standard approaches for NLI.\@ Due to space constraints, we give a brief description of the models here and refer the interested reader to the code repository for implementation details.

For experiments where premises are represented as sentences or paragraphs, we
evaluated a feature-based baseline using unigrams and bigrams of tokens. For
this model (referred to as \emph{\unibigrammodelname}), we used the LibLinear
library~\cite{fan2008liblinear}.

For these representations, we also evaluated a collection of BERT-class of
models. Following the standard setup, we encoded the premise-hypothesis pair,
and used the classification token to train a classifier, specifically a two-layer feedforward network that predicts the label. The hidden layer had half the size of the token embeddings. We compared RoBERTa$_{L}$ (Large), RoBERTa$_{B}$ (Base) and BERT$_{B}$ (Base) in our experiments. 

We used the above BERT strategy for the TabFact representations as well. For the
TabAttn representations, we implemented the popular decomposable attention
model~\cite{parikh-etal-2016-decomposable} using the premise key-value embeddings and hypothesis
token embeddings with 512 dimensional attend and compare layers.


We implemented all our models using the PyTorch with the transformers library~\cite{Wolf2019HuggingFacesTS}. We trained our models using Adagrad with a learning rate of $10^{-4}$, chosen by preliminary experiments, and using a dropout value of 0.2. All our results in the following sections are averages of models trained from three different random seeds. 

\subsection{Results}

Our experiments answer a series of questions.

\paragraph{Does our dataset exhibit hypothesis bias?} Before we consider the
question of whether we can model premise-hypothesis relationships, let us first
see if a model can learn to predict the entailment label without using the
premise, thereby exhibiting an undesirable artifact.  We consider three classes
of models to study hypothesis bias in \datasetName.

\noindent \textit{Hypothesis Only (hypo-only):} The simplest way to check for
hypothesis bias is to train a classifier using only the hypotheses. Without a premise, a classifier should fail to correlate the hypothesis and the
label. We represent the hypothesis in two ways a) using unigrams and bigrams for
an SVM, and b) using a single-sentence BERT-class model. The results of the experiments are given in Table \ref{tab:hypothesis_only_results}. 

\begin{table}[h]
\centering
\begin{tabular}{ccccc}
  \toprule
  Model & Dev & \alphaOne & \alphaTwo & \alphaThree \\
  \midrule
  Majority & 33.33 & 33.33 & 33.33 & 33.33 \\
  \unibigrammodelname &  59.00       & 60.61  & 45.89  & 45.89  \\
  BERT$_{B}$      &  62.69  & 63.45  & 49.65  &  50.45 \\ 
  RoBERTa$_{B}$      & 62.37   & 62.76  & 50.65  & 50.8  \\ 
  RoBERTa$_{L}$     & 60.51   &  60.48  & 48.26  &  48.89 \\
  \bottomrule
\end{tabular}
\caption{Accuracy of hypothesis-only baselines on the \datasetName Dev and test sets}
\label{tab:hypothesis_only_results}
\end{table}

\noindent \textit{Dummy or Swapped Premise:} Another approach to evaluate
hypothesis bias is to provide an unrelated premise and train a full entailment
model. We evaluated two cases, where every premise is changed to a
(a) \textit{dummy} statement (\example{to be or not to be}), or (b) a randomly
\textit{swapped} table that is represented as paragraph. In both cases, we
trained a RoBERTa$_L$ classifier as described in \S\ref{sec:models}. The results for these experiments are presented in Table \ref{tab:dummy_random_results}. 

\begin{table}[h]
\centering
\begin{tabular}{ccccc}
\toprule
Premise & Dev    & \alphaOne & \alphaTwo & \alphaThree \\ \midrule
dummy       & 60.02 & 59.78    & 48.91    & 46.37      \\
swapped     & 62.94  & 65.11     & 52.55     & 50.21      \\ 

\bottomrule
\end{tabular}
\caption{Accuracy with dummy/swapped premises}
\label{tab:dummy_random_results}
\end{table}

\noindent \textit{Results and Analysis: }
Looking at the Dev and \alphaOne columns of Tables
\ref{tab:hypothesis_only_results} and \ref{tab:dummy_random_results}, we see
that these splits do have hypothesis bias.  All the BERT-class models discover
such artifacts equally well. However, we also observe that the performance on
\alphaTwo~ and \alphaThree~ data splits is worse since the artifacts in the
training data do not occur in these splits. We see a performance gap of 
$\sim 12\%$ as compared to Dev and \alphaOne splits in all cases. While there
is some hypothesis bias in these splits, it is much less pronounced.

An important conclusion from these results is that the baseline for all future
models trained on these splits should be the best premise-free performance. From
the results here, these correspond to the \example{swapped} setting.

\skippedDetails{In Table \ref{tab:dummy_random_results}, we observe that providing the style and length of the premise helps the model predict on the original data. That is, we see a better accuracy for the model trained on swapped premise and tested on original premise when compared to the model trained with dummy premise and tested with the original premise.} 

\paragraph{How do trained NLI systems perform on our dataset?} Given the high
leaderboard accuracies of trained NLI systems, the question of whether these
models can infer entailment labels using a linearization of the tables
arises. To study this, we trained RoBERTa$_{L}$ models on the SNLI and MultiNLI
datasets. The SNLI model achieves an accuracy of 92.56$\%$ on SNLI test set. The
MultiNLI model achieves an accuracy of 89.0$\%$ on matched and 88.99$\%$ on the
mismatched MultiNLI test set. We evaluate these models on the WMD-1 and the Para
representations of premises.

\begin{table}[h] 
\centering
\begin{tabular}{cccccc}
\toprule
Premise & Dev   & \alphaOne & \alphaTwo & \bf \alphaThree       \\ \midrule 
\multicolumn{5}{c}{Trained on SNLI}        \\ 
WMD-1   & 49.44 & 47.5     & 49.44     & 46.44                 \\
Para    & 54.44 & 53.55     & 53.66     & 46.01                 \\\midrule
\multicolumn{5}{c}{Trained on MultiNLI } \\  
WMD-1   & 44.44 & 44.67     & 46.88     & 44.01                 \\
Para    & 55.77 & 53.83     & 55.33     & 47.28                 \\ \bottomrule
\end{tabular}
\caption{Accuracy of test splits with structured representation of premises with RoBERTa$_L$ trained on SNLI and MultiNLI training data}
\label{tab:snli_mnli_results}
\end{table}

\noindent \textit{Results and Analysis:} In Table \ref{tab:snli_mnli_results},
all the results point to the fact that pre-trained NLI systems do not perform
well when tested on
\datasetName. 
We observe that full premises slightly improve performance over the WMD-1
ones. This might be due to a) ineffectiveness of WMD to identify the correct
premise sentence, and b) multi-row reasoning.


\paragraph{Does training on the paragraph/sentence representation of a premise
  help?}
The next set of experiments compares BERT-class models and SVM trained using the paragraph (Para) and
sentence (WMD-n) representations.
The results for these experiments are presented in Table \ref{tab:baseline_results}.

\begin{table}[h]
\centering
\begin{tabular}{ccccc}
\toprule
Premise & Dev    & \alphaOne & \alphaTwo & \bf \alphaThree \\ \midrule 
\multicolumn{5}{c}{ Train with SVM }     \\
Para & 59.11 & 59.17 & 46.44 & 41.28 \\ \hline
\multicolumn{5}{c}{ Train with BERT$_{B}$}                 \\ 
Para    &  63.00    & 63.54     & 52.57     & 48.17           \\ \midrule
\multicolumn{5}{c}{ Train with RoBERTa$_{B}$}              \\ 
Para    & 67.2     & 66.98    & 56.87    & 55.36          \\ \midrule
\multicolumn{5}{c}{ Train with RoBERTa$_{L}$}              \\ 
WMD-1   & 65.44 & 65.27    & 57.11    & 52.55          \\
WMD-3   & 72.55 & 70.38    & 62.55   & 61.33          \\ 
Para    & 75.55     & 74.88    & 65.55    & 64.94          \\ \bottomrule
\end{tabular}
\caption{Accuracy of paragraph and sentence premise representation reported on SVM, BERT$_B$, RoBERTa$_B$ and RoBERTa$_L$}
\label{tab:baseline_results}
\end{table}

\noindent \textit{Results and Analysis: }
We find that training with the \datasetName training set improves model
performance significantly over the previous baselines, except for the simple SVM
model which relies on unigrams and bigrams. We see that  RoBERTa$_{L}$
outperforms its base variant and BERT$_B$ by around $\sim 9\%$ and $\sim 14\%$
respectively. Similar to the earlier observation, providing full premise is
better than selecting a subset of sentences.

Importantly, \alphaTwo and \alphaThree performance is worse than \alphaOne, not
only suggesting the difficulty of these data splits, but also showing that
models overfit both lexical patterns (based on \alphaTwo) or domain-specific
patterns (based on \alphaThree).

\paragraph{Does training on premise encoded as structure help?}
Rather than linearizing the tables as sentences, we can try to encode the structure of the tables. We consider two representative approaches for this, TabFact and TabAttn, each associated with a different model as described in \S\ref{sec:models}.
The results for these experiments are listed in Table \ref{tab:structure_results}.

\begin{table}[h]
\centering
\begin{tabular}{ccccc}
\toprule
Premise & Dev   & \alphaOne & \alphaTwo & \alphaThree \\ \midrule
\multicolumn{5}{c}{ Train with BERT$_{B}$}            \\
TabFact & 63.67 & 64.04     & 53.59     & 49.05       \\ \midrule
\multicolumn{5}{c}{ Train with RoBERT$_{B}$}          \\
TabFact & 68.06 & 66.7      & 56.87     & 55.26       \\ \midrule
\multicolumn{5}{c}{ Train with RoBERTa$_{L}$}         \\ 
TabAttn & 63.63 & 62.94     & 49.37     & 49.04       \\ 
TabFact & 77.61 & 75.06      & 69.02     & 64.61       \\    \bottomrule  
\end{tabular}

\caption{Accuracy on structured premise representation reported on BERT$_B$, RoBERTa$_B$ and RoBERTa$_L$}
\label{tab:structure_results}
\end{table}

\noindent \textit{Results and Analysis: }
The idea of using this family of models was to leverage the structural aspects of our data. We find that the TabAttn model, however, does not improve the performance. We assume that this might be due to the bag of words style of representation that the classifier employs.  We find, however, that providing premise structure information helps the TabFact model perform better than the RoBERTa$_{L}$+Para model. As before model performance drops for \alphaTwo~and \alphaThree.

\paragraph{How many types of reasoning does a trained system predict correctly?}

\begin{figure*}
  \centering
  \begin{subfigure}[b]{0.44\textwidth}
    \centering
    \includegraphics[width=\textwidth]{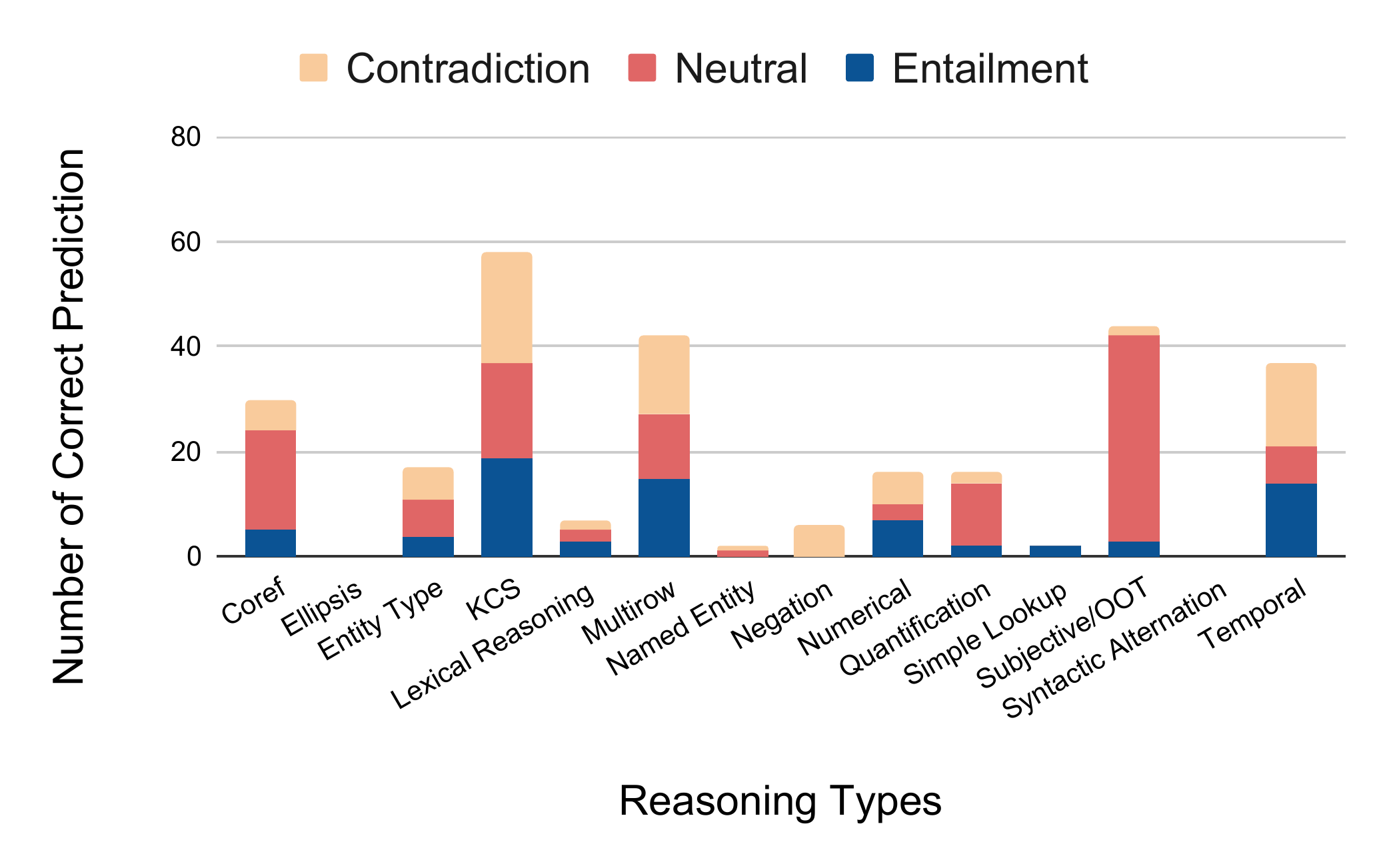}
    \caption{Number of correct predictions per reasoning type in the Dev set}
    \label{fig:pred1_dev}
  \end{subfigure}
  \hfill
  \begin{subfigure}[b]{0.44\textwidth}  
    \centering 
    \includegraphics[width=\textwidth]{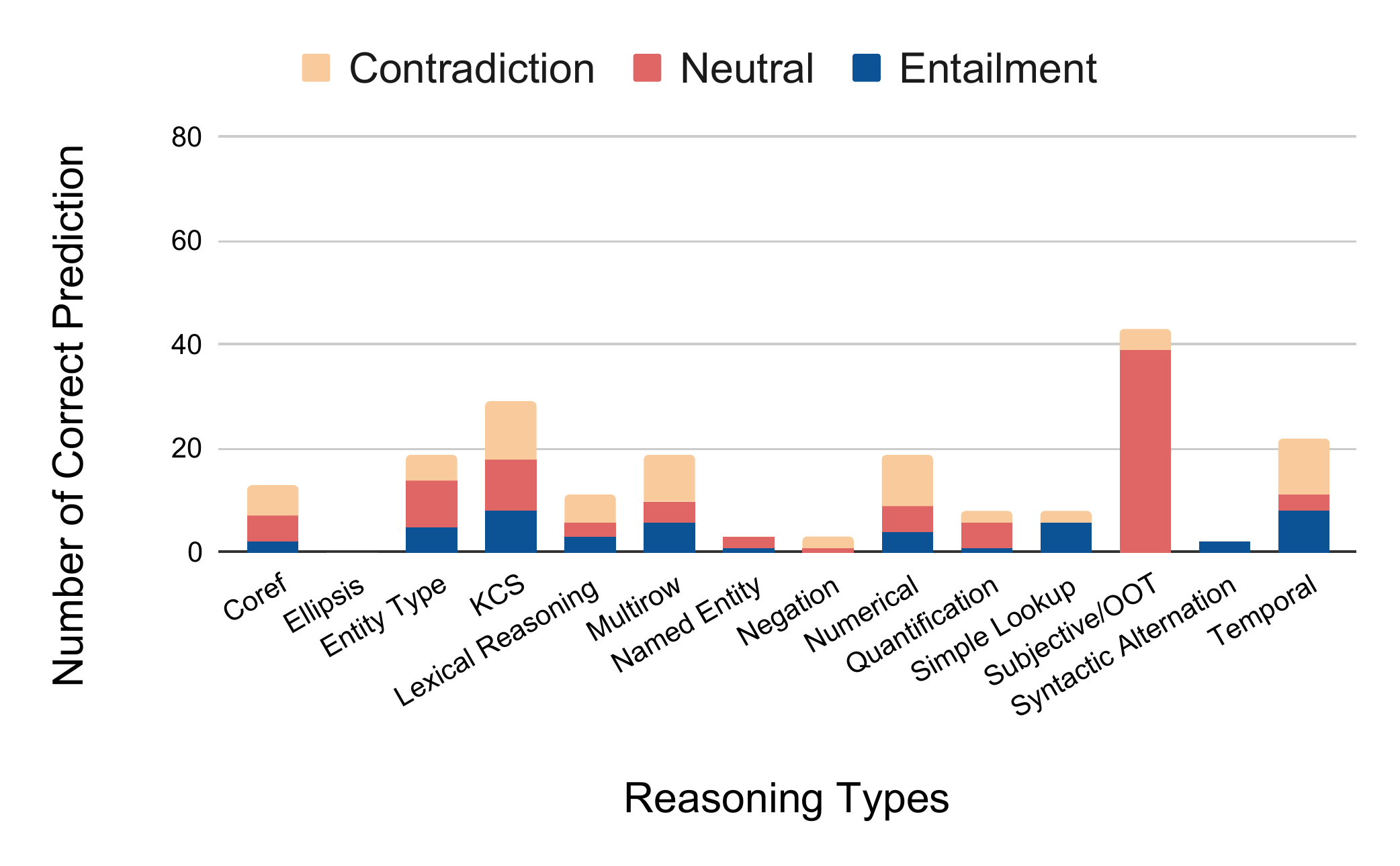}
    \caption{Number of correct predictions per reasoning type in the~\alphaThree test set}
    \label{fig:pred1_alphaThree}
  \end{subfigure}
  \caption{Number of correct predictions per reasoning type in the Dev and \alphaThree splits.} 
  \label{fig: reasoning prediction statistics}
\end{figure*}

Using a RoBERTa$_{L}$, which was trained on the paragraph (Para) representation, we analyzed the examples in Dev and ~\alphaThree data splits that were annotated by experts for their types of reasoning (\S\ref{sec:main-reasoning}). Figure~\ref{fig: reasoning prediction statistics} shows the summary of this analysis.

\noindent \textit{Results and Analysis: } Figures \ref{fig:pred1_dev} and \ref{fig:pred1_alphaThree} show the histogram of reasoning types among correctly predicted examples. Compared to Figures~\ref{fig:reasoning_dev} and \ref{fig:reasoning_alphaThree}, we see a decrease in correct predictions across all reasoning types for both Dev and~\alphaThree sets.  In particular, in the Dev set, the model performs poorly for the knowledge \& common sense, multi-row, coreference, and temporal reasoning categories. 

\paragraph{Discussion} Our results show that:
\begin{inparaenum}[1)]
\item \datasetName contains a certain amount of artifacts which transformer-based
  models learn, but all models have a large gap to human performance; and
\item models accuracies drop on \alphaTwo~and \alphaThree, suggesting that all three results together should be used to characterize the model, and not any single one of them.
\end{inparaenum}
All our models are significantly worse than the human performance ($84.04\%$,
$83.88\%$ and $79.33\%$ for \alphaOne, \alphaTwo and \alphaThree respectively).
With a difference of $\sim 14\%$ between our best model and the human performance, these results indicate that \datasetName is a challenging dataset.


\section{Related Work}
\label{sec:related}

\paragraph{NLI Datasets} Natural language inference/textual entailment is a
well studied text understanding task, and has several datasets of various
sizes. The annual PASCAL RTE challenges~\cite[][{\em inter alia}]{dagan2005pascal} were
associated with several thousands of human-annotated entailment pairs. The SNLI
dataset~\cite{snli:emnlp2015} is the first large scale entailment dataset that
uses image captions as premises, while the MultiNLI~\cite{N18-1101} uses
premises from multiple domains.
The QNLI and WNLI datasets provide a new perspective by converting the SQuAD
question answering data~\cite{rajpurkar-etal-2016-squad} and Winograd Schema
Challenge data~\citep{levesque2012winograd} respectively into inference
tasks. More recently, SciTail~\cite{scitail} and Adversarial
NLI~\cite{nie2019adversarial} have focused on building adversarial datasets; the
former uses information retrieval to select adversarial premises, while the
latter uses iterative annotation cycles to confuse models.

\paragraph{Reasoning} Recently, challenging new datasets have emerged that emphasize complex reasoning. \citet{bhagavatula2019abductive} pose the task of determining the most plausible inferences based on observation (abductive reasoning). Across NLP, a lot of work has been published around different kinds of reasonings. To name a few, common sense~\cite{talmor-etal-2019-commonsenseqa}, temporal~\cite{zhou-etal-2019-going}, numerical~\cite{naik-etal-2019-exploring, wallace2019nlp} and multi-hop~\cite{khashabi2018looking} reasoning have all garnered immense research interest.

\paragraph{Tables and Semi-structured data}
Tasks based on semi-structured data in the form of tables, graphs and databases (with entries as text) contain complex reasoning~\cite{dhingra-etal-2019-handling, 2019TabFactA}. Previous work has touched upon semantic parsing and question answering~\cite[e.g.,][and references therein]{pasupat2015compositional,khashabi2016question}, which typically work with tables with many entries that resemble database records.

Our work is most closely related to TabFact~\cite{2019TabFactA}, which considers database-style tables as premises with human-annotated hypotheses to form an inference task. While there are similarities in the task formulation scheme, our work presents an orthogonal perspective:
\begin{inparaenum}[(i)]
\item The Wikipedia tables premises of TabFact are homogeneous, i.e., each column in a table has structural redundancy and all entries have the same type. One can look at multiple entries of a column to infer extra information, e.g., all entries of a column are about locations. On the contrary, the premises in our dataset are heterogeneous.
\item TabFact only considers entailment and contradiction; we argue that inference is non-binary with a third ``undetermined'' class (neutrals).  
\item Compared to our multi-faceted reasonings, the reasonings of the hypotheses in TabFact are limited and mostly numerical or comparatives.
\item The  \alphaTwo~and \alphaThree sets help us check for annotation and domain-specific artifacts.
\end{inparaenum}

\paragraph{Artifacts} 
Recently, pre-trained transformer-based models \cite[][and others]{devlin2019bert,radford2019language,liu2019roberta} have seemingly outperformed human performance on several NLI tasks~\cite{wang2018glue,wang2019superglue}. However, it has been shown by \citet{poliak-etal-2018-hypothesis,niven-kao-2019-probing,gururangan2018annotation,glockner2018breaking,naik2018stress,wallace2019universal} that these models exploit spurious patterns (artifacts) in the data to obtain good performance. It is imperative to produce datasets that allow for controlled study of artifacts. A popular strategy today is to use adversarial annotation \cite{zellers2018swagaf,nie2019adversarial} and rewriting of the input \cite{2019TabFactA}.  We argue that we can systematically construct test sets that can help study artifacts along specific dimensions.


\section{Conclusion}
\label{sec:conclusion}

We presented a new high quality natural language inference dataset,
\datasetName, with heterogeneous semi-structured premises and natural language
hypotheses. Our analysis showed that our data encompasses several different
kinds of inferences.  \datasetName has multiple test sets that
are designed to pose difficulties to models that only learn superficial
correlations between inputs and the labels, rather than reasoning about the
information.
Via extensive experiments, we showed that derivatives of several popular classes
of models find this new inference task challenging. We expect that the dataset
can serve as a testbed for developing new kinds of models and representations
that can handle semi-structured information as first class citizens.


\section*{Acknowledgements}
\label{sec:aknowledgement}

We thank members of the Utah NLP group for their valuable insights and
suggestions at various stages of the project; and reviewers their helpful comments. We acknowledge the support
of the support of NSF Grants No. 1822877 and 1801446, and a generous gift from
Google.


\bibliographystyle{acl_natbib.bst}
\bibliography{cited}

\begin{thebibliography}{34}
\expandafter\ifx\csname natexlab\endcsname\relax\def\natexlab#1{#1}\fi

\bibitem[{Artstein and Poesio(2008)}]{artstein2008inter}
Ron Artstein and Massimo Poesio. 2008.
\newblock \href {https://doi.org/10.1162/coli.07-034-R2} {{Inter-coder
  Agreement for Computational Linguistics}}.
\newblock \emph{Computational Linguistics}, 34(4):555--596.

\bibitem[{Bhagavatula et~al.(2020)Bhagavatula, Bras, Malaviya, Sakaguchi,
  Holtzman, Rashkin, Downey, Yih, and Choi}]{bhagavatula2019abductive}
Chandra Bhagavatula, Ronan~Le Bras, Chaitanya Malaviya, Keisuke Sakaguchi, Ari
  Holtzman, Hannah Rashkin, Doug Downey, Scott Wen-tau Yih, and Yejin Choi.
  2020.
\newblock \href {https://openreview.net/forum?id=Byg1v1HKDB} {{Abductive
  Commonsense Reasoning}}.
\newblock In \emph{International Conference on Learning Representations}.

\bibitem[{Bowman et~al.(2015)Bowman, Angeli, Potts, and
  Manning}]{snli:emnlp2015}
Samuel~R. Bowman, Gabor Angeli, Christopher Potts, and Christopher~D. Manning.
  2015.
\newblock \href {https://doi.org/10.18653/v1/D15-1075} {{A Large Annotated
  Corpus for Learning Natural Language Inference}}.
\newblock In \emph{Proceedings of the 2015 Conference on Empirical Methods in
  Natural Language Processing}.

\bibitem[{Chen et~al.(2020)Chen, Wang, Chen, Zhang, Wang, Li, Zhou, and
  Wang}]{2019TabFactA}
Wenhu Chen, Hongmin Wang, Jianshu Chen, Yunkai Zhang, Hong Wang, Shiyang Li,
  Xiyou Zhou, and William~Yang Wang. 2020.
\newblock \href {https://openreview.net/forum?id=rkeJRhNYDH} {{TabFact : A
  Large-scale Dataset for Table-based Fact Verification}}.
\newblock In \emph{International Conference on Learning Representations}.

\bibitem[{Dagan et~al.(2005)Dagan, Glickman, and Magnini}]{dagan2005pascal}
Ido Dagan, Oren Glickman, and Bernardo Magnini. 2005.
\newblock \href {https://doi.org/10.1007/11736790_9} {{The PASCAL Recognising
  Textual Entailment Challenge}}.
\newblock In \emph{Machine Learning Challenges Workshop}, pages 177--190.
  Springer.

\bibitem[{Devlin et~al.(2019)Devlin, Chang, Lee, and
  Toutanova}]{devlin2019bert}
Jacob Devlin, Ming-Wei Chang, Kenton Lee, and Kristina Toutanova. 2019.
\newblock \href {https://doi.org/10.18653/v1/N19-1423} {{BERT: Pre-training of
  Deep Bidirectional Transformers for Language Understanding}}.
\newblock In \emph{Proceedings of the 2019 Conference of the North American
  Chapter of the Association for Computational Linguistics: Human Language
  Technologies}.

\bibitem[{Dhingra et~al.(2019)Dhingra, Faruqui, Parikh, Chang, Das, and
  Cohen}]{dhingra-etal-2019-handling}
Bhuwan Dhingra, Manaal Faruqui, Ankur Parikh, Ming-Wei Chang, Dipanjan Das, and
  William Cohen. 2019.
\newblock \href {https://doi.org/10.18653/v1/P19-1483} {{Handling Divergent
  Reference Texts when Evaluating Table-to-Text Generation}}.
\newblock In \emph{Proceedings of the 57th Annual Meeting of the Association
  for Computational Linguistics}.

\bibitem[{Fan et~al.(2008)Fan, Chang, Hsieh, Wang, and Lin}]{fan2008liblinear}
Rong-En Fan, Kai-Wei Chang, Cho-Jui Hsieh, Xiang-Rui Wang, and Chih-Jen Lin.
  2008.
\newblock \href
  {https://dl.acm.org/doi/pdf/10.5555/1390681.1442794?download=true}
  {{LIBLINEAR: A Library for Large Linear Classification}}.
\newblock \emph{Journal of Machine Learning Research}.

\bibitem[{Glockner et~al.(2018)Glockner, Shwartz, and
  Goldberg}]{glockner2018breaking}
Max Glockner, Vered Shwartz, and Yoav Goldberg. 2018.
\newblock \href {https://doi.org/10.18653/v1/P18-2103} {{Breaking NLI Systems
  with Sentences that Require Simple Lexical Inferences}}.
\newblock In \emph{Proceedings of the 56th Annual Meeting of the Association
  for Computational Linguistics}.

\bibitem[{Gururangan et~al.(2018)Gururangan, Swayamdipta, Levy, Schwartz,
  Bowman, and Smith}]{gururangan2018annotation}
Suchin Gururangan, Swabha Swayamdipta, Omer Levy, Roy Schwartz, Samuel Bowman,
  and Noah~A Smith. 2018.
\newblock \href {https://doi.org/10.18653/v1/N18-2017} {{Annotation Artifacts
  in Natural Language Inference Data}}.
\newblock In \emph{Proceedings of the 2018 Conference of the North American
  Chapter of the Association for Computational Linguistics: Human Language
  Technologies}.

\bibitem[{Khashabi et~al.(2018)Khashabi, Chaturvedi, Roth, Upadhyay, and
  Roth}]{khashabi2018looking}
Daniel Khashabi, Snigdha Chaturvedi, Michael Roth, Shyam Upadhyay, and Dan
  Roth. 2018.
\newblock \href {https://doi.org/10.18653/v1/N18-1023} {{Looking Beyond the
  Surface: A Challenge Set for Reading Comprehension over Multiple Sentences}}.
\newblock In \emph{Proceedings of the 2018 Conference of the North American
  Chapter of the Association for Computational Linguistics: Human Language
  Technologies}.

\bibitem[{Khashabi et~al.(2016)Khashabi, Khot, Sabharwal, Clark, Etzioni, and
  Roth}]{khashabi2016question}
Daniel Khashabi, Tushar Khot, Ashish Sabharwal, Peter Clark, Oren Etzioni, and
  Dan Roth. 2016.
\newblock \href {http://www.ijcai.org/Proceedings/16/Papers/166.pdf} {{Question
  Answering via Integer Programming over Semi-structured Knowledge}}.
\newblock In \emph{Proceedings of the Twenty-Fifth International Joint
  Conference on Artificial Intelligence}.

\bibitem[{Khot et~al.(2018)Khot, Sabharwal, and Clark}]{scitail}
Tushar Khot, Ashish Sabharwal, and Peter Clark. 2018.
\newblock \href
  {https://www.aaai.org/ocs/index.php/AAAI/AAAI18/paper/view/17368/16067}
  {{{SciTail}: A Textual Entailment Dataset from Science Question Answering}}.
\newblock In \emph{Association for the Advancement of Artificial Intelligence}.

\bibitem[{Kusner et~al.(2015)Kusner, Sun, Kolkin, and
  Weinberger}]{kusner2015word}
Matt Kusner, Yu~Sun, Nicholas Kolkin, and Kilian Weinberger. 2015.
\newblock \href {http://jmlr.org/proceedings/papers/v37/kusnerb15.pdf} {{From
  Word Embeddings to Document Distances}}.
\newblock In \emph{International Conference on Machine Learning}.

\bibitem[{Levesque et~al.(2012)Levesque, Davis, and
  Morgenstern}]{levesque2012winograd}
Hector Levesque, Ernest Davis, and Leora Morgenstern. 2012.
\newblock \href
  {https://www.aaai.org/ocs/index.php/KR/KR12/paper/view/4492/4924} {{The
  Winograd Schema Challenge}}.
\newblock In \emph{Thirteenth International Conference on the Principles of
  Knowledge Representation and Reasoning}.

\bibitem[{Liu et~al.(2019)Liu, Ott, Goyal, Du, Joshi, Chen, Levy, Lewis,
  Zettlemoyer, and Stoyanov}]{liu2019roberta}
Yinhan Liu, Myle Ott, Naman Goyal, Jingfei Du, Mandar Joshi, Danqi Chen, Omer
  Levy, Mike Lewis, Luke Zettlemoyer, and Veselin Stoyanov. 2019.
\newblock \href {https://openreview.net/forum?id=SyxS0T4tvS} {{Roberta: A
  Robustly Optimized BERT Pretraining Approach}}.
\newblock \emph{arXiv preprint arXiv:1907.11692}.

\bibitem[{Naik et~al.(2019)Naik, Ravichander, Rose, and
  Hovy}]{naik-etal-2019-exploring}
Aakanksha Naik, Abhilasha Ravichander, Carolyn Rose, and Eduard Hovy. 2019.
\newblock \href {https://doi.org/10.18653/v1/P19-1329} {{Exploring Numeracy in
  Word Embeddings}}.
\newblock In \emph{Proceedings of the 57th Annual Meeting of the Association
  for Computational Linguistics}.

\bibitem[{Naik et~al.(2018)Naik, Ravichander, Sadeh, Rose, and
  Neubig}]{naik2018stress}
Aakanksha Naik, Abhilasha Ravichander, Norman Sadeh, Carolyn Rose, and Graham
  Neubig. 2018.
\newblock \href {https://www.aclweb.org/anthology/C18-1198} {{Stress Test
  Evaluation for Natural Language Inference}}.
\newblock In \emph{Proceedings of the 27th International Conference on
  Computational Linguistics}.

\bibitem[{Nie et~al.(2019)Nie, Williams, Dinan, Bansal, Weston, and
  Kiela}]{nie2019adversarial}
Yixin Nie, Adina Williams, Emily Dinan, Mohit Bansal, Jason Weston, and Douwe
  Kiela. 2019.
\newblock \href {https://arxiv.org/pdf/1910.14599.pdf} {{Adversarial NLI: A New
  Benchmark for Natural Language Understanding}}.
\newblock \emph{arXiv preprint arXiv:1910.14599}.

\bibitem[{Niven and Kao(2019)}]{niven-kao-2019-probing}
Timothy Niven and Hung-Yu Kao. 2019.
\newblock \href {https://doi.org/10.18653/v1/P19-1459} {{Probing Neural Network
  Comprehension of Natural Language Arguments}}.
\newblock In \emph{Proceedings of the 57th Annual Meeting of the Association
  for Computational Linguistics}.

\bibitem[{Parikh et~al.(2016)Parikh, T{\"a}ckstr{\"o}m, Das, and
  Uszkoreit}]{parikh-etal-2016-decomposable}
Ankur Parikh, Oscar T{\"a}ckstr{\"o}m, Dipanjan Das, and Jakob Uszkoreit. 2016.
\newblock \href {https://doi.org/10.18653/v1/D16-1244} {{A Decomposable
  Attention Model for Natural Language Inference}}.
\newblock In \emph{Proceedings of the 2016 Conference on Empirical Methods in
  Natural Language Processing}.

\bibitem[{Pasupat and Liang(2015)}]{pasupat2015compositional}
Panupong Pasupat and Percy Liang. 2015.
\newblock \href {https://doi.org/10.3115/v1/P15-1142} {{Compositional Semantic
  Parsing on Semi-Structured Tables}}.
\newblock In \emph{Proceedings of the 53rd Annual Meeting of the Association
  for Computational Linguistics and the 7th International Joint Conference on
  Natural Language Processing}.

\bibitem[{Poliak et~al.(2018)Poliak, Naradowsky, Haldar, Rudinger, and
  Van~Durme}]{poliak-etal-2018-hypothesis}
Adam Poliak, Jason Naradowsky, Aparajita Haldar, Rachel Rudinger, and Benjamin
  Van~Durme. 2018.
\newblock \href {https://doi.org/10.18653/v1/S18-2023} {{Hypothesis Only
  Baselines in Natural Language Inference}}.
\newblock In \emph{Proceedings of the Seventh Joint Conference on Lexical and
  Computational Semantics}.

\bibitem[{Radford et~al.(2019)Radford, Wu, Child, Luan, Amodei, and
  Sutskever}]{radford2019language}
Alec Radford, Jeffrey Wu, Rewon Child, David Luan, Dario Amodei, and Ilya
  Sutskever. 2019.
\newblock \href
  {https://www.ceid.upatras.gr/webpages/faculty/zaro/teaching/alg-ds/PRESENTATIONS/PAPERS/2019-Radford-et-al_Language-Models-Are-Unsupervised-Multitask-\%20Learners.pdf}
  {{Language Models are Unsupervised Multitask Learners}}.
\newblock In \emph{OpenAI Blog}.

\bibitem[{Rajpurkar et~al.(2016)Rajpurkar, Zhang, Lopyrev, and
  Liang}]{rajpurkar-etal-2016-squad}
Pranav Rajpurkar, Jian Zhang, Konstantin Lopyrev, and Percy Liang. 2016.
\newblock \href {https://doi.org/10.18653/v1/D16-1264} {{{SQ}u{AD}: 100,000+
  Questions for Machine Comprehension of Text}}.
\newblock In \emph{Proceedings of the 2016 Conference on Empirical Methods in
  Natural Language Processing}.

\bibitem[{Talmor et~al.(2019)Talmor, Herzig, Lourie, and
  Berant}]{talmor-etal-2019-commonsenseqa}
Alon Talmor, Jonathan Herzig, Nicholas Lourie, and Jonathan Berant. 2019.
\newblock \href {https://doi.org/10.18653/v1/N19-1421} {{{C}ommonsense{QA}: A
  Question Answering Challenge Targeting Commonsense Knowledge}}.
\newblock In \emph{Proceedings of the 2019 Conference of the North {A}merican
  Chapter of the Association for Computational Linguistics: Human Language
  Technologies}, Minneapolis, Minnesota.

\bibitem[{Wallace et~al.(2019{\natexlab{a}})Wallace, Feng, Kandpal, Gardner,
  and Singh}]{wallace2019universal}
Eric Wallace, Shi Feng, Nikhil Kandpal, Matt Gardner, and Sameer Singh.
  2019{\natexlab{a}}.
\newblock \href {https://doi.org/10.18653/v1/D19-1221} {{Universal Adversarial
  Triggers for Attacking and Analyzing NLP}}.
\newblock In \emph{Proceedings of the 2019 Conference on Empirical Methods in
  Natural Language Processing and the 9th International Joint Conference on
  Natural Language Processing}.

\bibitem[{Wallace et~al.(2019{\natexlab{b}})Wallace, Wang, Li, Singh, and
  Gardner}]{wallace2019nlp}
Eric Wallace, Yizhong Wang, Sujian Li, Sameer Singh, and Matt Gardner.
  2019{\natexlab{b}}.
\newblock \href {https://doi.org/10.18653/v1/D19-1534} {{Do NLP Models Know
  Numbers? Probing Numeracy in Embeddings}}.
\newblock In \emph{Proceedings of the 2019 Conference on Empirical Methods in
  Natural Language Processing and the 9th International Joint Conference on
  Natural Language Processing}.

\bibitem[{Wang et~al.(2019{\natexlab{a}})Wang, Pruksachatkun, Nangia, Singh,
  Michael, Hill, Levy, and Bowman}]{wang2019superglue}
Alex Wang, Yada Pruksachatkun, Nikita Nangia, Amanpreet Singh, Julian Michael,
  Felix Hill, Omer Levy, and Samuel~R Bowman. 2019{\natexlab{a}}.
\newblock \href
  {http://papers.nips.cc/paper/8589-superglue-a-stickier-benchmark-for-general-purpose-language-understanding-systems.pdf}
  {{SuperGLUE: A Stickier Benchmark for General-Purpose Language Understanding
  Systems}}.
\newblock In \emph{Advances in Neural Information Processing Systems}.

\bibitem[{Wang et~al.(2019{\natexlab{b}})Wang, Singh, Michael, Hill, Levy, and
  Bowman}]{wang2018glue}
Alex Wang, Amanpreet Singh, Julian Michael, Felix Hill, Omer Levy, and Samuel~R
  Bowman. 2019{\natexlab{b}}.
\newblock \href {https://doi.org/10.18653/v1/W18-5446} {{GLUE: A Multi-task
  Benchmark and Analysis Platform for Natural Language Understanding}}.
\newblock In \emph{International Conference on Learning Representations}.

\bibitem[{Williams et~al.(2018)Williams, Nangia, and Bowman}]{N18-1101}
Adina Williams, Nikita Nangia, and Samuel Bowman. 2018.
\newblock \href {https://doi.org/10.18653/v1/N18-1101} {{A Broad-Coverage
  Challenge Corpus for Sentence Understanding through Inference}}.
\newblock In \emph{Proceedings of the 2018 Conference of the North American
  Chapter of the Association for Computational Linguistics: Human Language
  Technologies}.

\bibitem[{Wolf et~al.(2019)Wolf, Debut, Sanh, Chaumond, Delangue, Moi, Cistac,
  Rault, Louf, Funtowicz, and Brew}]{Wolf2019HuggingFacesTS}
Thomas Wolf, Lysandre Debut, Victor Sanh, Julien Chaumond, Clement Delangue,
  Anthony Moi, Pierric Cistac, Tim Rault, R'emi Louf, Morgan Funtowicz, and
  Jamie Brew. 2019.
\newblock \href {https://arxiv.org/pdf/1910.03771.pdf} {{HuggingFace's
  Transformers: State-of-the-art Natural Language Processing}}.
\newblock \emph{ArXiv}.

\bibitem[{Zellers et~al.(2018)Zellers, Bisk, Schwartz, and
  Choi}]{zellers2018swagaf}
Rowan Zellers, Yonatan Bisk, Roy Schwartz, and Yejin Choi. 2018.
\newblock \href {https://doi.org/10.18653/v1/D18-1009} {{SWAG: A Large-Scale
  Adversarial Dataset for Grounded Commonsense Inference}}.
\newblock In \emph{Proceedings of the 2018 Conference on Empirical Methods in
  Natural Language Processing}.

\bibitem[{Zhou et~al.(2019)Zhou, Khashabi, Ning, and
  Roth}]{zhou-etal-2019-going}
Ben Zhou, Daniel Khashabi, Qiang Ning, and Dan Roth. 2019.
\newblock \href {https://doi.org/10.18653/v1/D19-1332} {{{``}Going on a
  vacation{''} takes longer than {``}Going for a walk{''}: A Study of Temporal
  Commonsense Understanding}}.
\newblock In \emph{Proceedings of the 2019 Conference on Empirical Methods in
  Natural Language Processing and the 9th International Joint Conference on
  Natural Language Processing}.

\end{thebibliography}

\appendix

\section{Examples of Data}
\label{sec:data_examples}

Figure~\ref{fig:example_of_data} shows two additional examples of table premises
and their corresponding hypotheses available in the development set of
\datasetName. 

\begin{figure}[h!]
  \centering
  {
  \footnotesize
  \begin{center}
    \begin{tabular}{>{\raggedright}p{0.3\linewidth}p{0.45\linewidth}}
      \multicolumn{2}{c}{\bf Kamloops}   
                        \\
      \toprule
                                      
      {\bf Type}                & Elected city council                                                  \\
      {\bf Mayor}           & Ken Christian         \\
      {\bf Governing body}           & Kamloops City Council                                               \\
      {\bf MP}              & Cathy McLeod                                \\
      {\bf MLAs}                  &Peter Milobar, Todd Stone                                                                  \\
      \bottomrule
    \end{tabular}
  \end{center}
}
  {\footnotesize
    \begin{enumerate}[nosep]
    \item[H1:] Kamloops has a democracy structure.
    \item[H2:] If Ken Christian resigns as Mayor of Kamloops then Cathy McLeod will most likely replace him.
    \item[H3:] Kamloops is ruled by a president.
    
    \end{enumerate}
  }
\end{figure}
\begin{figure}[h!]
  \centering
  {
  \footnotesize
  \begin{center}
    \begin{tabular}{>{\raggedright}p{0.3\linewidth}p{0.45\linewidth}}
      \multicolumn{2}{c}{\bf Jefferson Starship}   
                        \\
      \toprule
                                      
      {\bf Origin}                & San Francisco California                                                  \\
      {\bf Genres}           & Rock, hard rock, psychedelic rock, progressive rock, soft rock      \\
      {\bf Years active}              & 1970 - 1984, 1992 - present                              \\
      {\bf Labels}                  &  RCA Grunt Epic\\
       {\bf Associated acts}              & Jefferson Airplane Starship, KBC Band, Hot Tuna  \\
       {\bf Website}              & www.jeffersonstarship.net\\                           
      \bottomrule
    \end{tabular}
  \end{center}
}
  {\footnotesize
    \begin{enumerate}[nosep]
    \item[H1:] Jefferson Starship was started on the West Coast of the United States.
    \item[H2:] Jefferson Starship won many awards for its music.
    \item[H3:] Jefferson Starship has performed continuously since the 1970s.
    
    \end{enumerate}
  }
  \caption{Two semi-structured premises (the tables), and three hypotheses (H1: entailment, H2: Neutral, and H3: contradiction) that correspond to each table.}
  \label{fig:example_of_data}
\end{figure}


\section{Reasoning for \datasetName}
\label{sec:reasoning}

Our inventory of reasoning types is based on GLUE diagnostics~\cite{wang2018glue}, but is specialized to the problem of reasoning about tables. Consequently, some categories from GLUE diagnostics may not be represented here, or may be merged into one category.

We assume that the table is correct and complete. The former is always true for textual entailment, where we assume that the premise is correct. The latter need not be generally true. However, in our analysis, we assume that the table lists all the relevant information for a field. For example, in a table for a music group as in Figure~\ref{fig:example_of_data}, if there is a row called \textbf{Labels}, we will assume that the labels listed in that row are the only labels associated with the group.

Note that a single premise-hypothesis pair may be associated with multiple types of reasoning. If the same reasoning type is employed multiple times in the same pair, we only mark it once.

\paragraph{Simple lookup} This is the simple case where there is no reasoning, and the hypothesis is formed by literally restating information in the table. For example, using the table in Figure~\ref{fig:example_picasso}, \example{Femme aux Bras Crois\'es is privately held.} is a simple lookup.

\paragraph{Multi-row reasoning}
Multiple rows in the table are needed to make an inference. This has the strong requirement that without multiple rows, there is no way to arrive at the conclusion.
Exclude instances where multiple rows are used only to identify the type of the entity, which is then used to make an inference. The test for multi-row reasoning is: If a row is removed from the table, then the label for the hypothesis may change.

\paragraph{Entity type}
Involves ascertaining the type of an entity in question (perhaps using multiple rows from the table), and then using this information to make an inference about the entity.

This is separate from multi-row reasoning even if discovering the entity type might require reading multiple rows in the table.
The difference is a practical one: we want to identify how many inferences in the data require multiple rows (both keys and values) separately from the ones that just use information about the entity type.
We need to be able to identify an entity and its type separately to decide on this category. In addition, while multi-row reasoning, by definition, needs multiple rows, entity type may be determined by looking at one row. For instance, looking at Figure \ref{fig:example_picasso}, one can infer that the entity type is a \example{painting} by only looking at the row with key value \textbf{Medium}. Lastly, ascertaining the entity type may require knowledge, but if so, then we will not explicitly mark the instance as Knowledge \& Common Sense.
For example, knowing that SNL is a TV show will be entity type and not Knowledge \& Common Sense.

\paragraph{Lexical reasoning}
Any inference that can be made using words, independent of the context of the words falls. For example, knowing that dogs are animals, and alive contradicts dead would fall into the category of lexical reasoning. This type of reasoning includes substituting words with their synonyms, hypernyms, hyponyms and antonyms. It also includes cases where  a semantically equivalent or contradicting word (perhaps belonging to a different root word) is used in the hypothesis., e.g., replacing understand with miscomprehend. 
Lexical reasoning also includes reasoning about monotonicity of phrases.

\paragraph{Negation}
Any explicit negation, including morphological negation (e.g., the word \example{affected} being mapped to \example{unaffected}).
Negation changes the morphology without changing the root word, e.g., we have to add an explicit \example{not}. 

This category includes double negations, which we believe is rare in our data. For example, the introduction of the phrase \example{not impossible} would count as a double negation.
If the word \example{understand} in the premise is replaced with \example{not comprehend}, we are changing the root word (understand to comprehend) and introducing a negation. So this change will be marked as both Lexical reasoning and Negation.

\begin{figure}[t]
  \centering
  {
  \footnotesize
  \begin{center}
    \begin{tabular}{>{\raggedright}p{0.3\linewidth}p{0.45\linewidth}}
      \multicolumn{2}{c}{\bf Femme aux Bras Crois\'es}   
                        \\
      \toprule
                                      
      {\bf Artist}                & Pablo Picasso                                                   \\
      {\bf Year}           & 1901-02          \\
      {\bf Medium}           & Oil on canvas                                                 \\
      {\bf Dimensions}              & 81 cm 58 cm (32 in 23 in)                                  \\
      {\bf Location}                  & Privately held                                                                 \\
      \bottomrule
    \end{tabular}
  \end{center}
}
  {\footnotesize
  }
  \caption{An example premise}
  \label{fig:example_picasso}
\end{figure}


\paragraph{Knowledge $\&$ Common Sense}
This category is related to the World Knowledge and Common Sense categories from GLUE. To quote the description from GLUE: ``...the entailment rests not only on correct disambiguation of the sentences, but also application of extra knowledge, whether it is concrete knowledge about world affairs or more common-sense knowledge about word meanings or social or physical dynamics.''

While GLUE differentiates between world knowledge and common sense, we found that this distinction is not always clear when reasoning about tables. So we do not make the distinction.

\paragraph{Named Entities}
This category is identical to the Named Entities category from GLUE. It includes an understanding of the compositional aspect of names (for example, knowing that the \example{University of Hogwarts} is the same as {Hogwarts}). Acronyms and their expansions fall into this category (e.g., the equivalence of \example{New York Stock Exchange} as \example{NYSE}).

\paragraph{Numerical reasoning}
Any form of reasoning that involves understanding numbers, counting, ranking, intervals and units falls under this group. This category also includes numerical comparisons and the use of mathematical operators to arrive at the hypothesis.

\paragraph{Temporal reasoning}
Any inferences that involves reasoning about time fall into this category. There may be an overlap between other categories and this one. Any numerical reasoning about temporal quantities and the use of knowledge about time should be included here. Examples of temporal reasoning:
\begin{itemize}
    \item 9 AM is in the morning. (Since this is knowledge about time, we will only tag this as Temporal.)
    \item 1950 is the $20^{th}$ century. 
    \item 1950 to 1962 is twelve years.
    \item Steven Spielberg was born in the winter of 1946. (If the table has the date---18th December, 1946---and the location of birth---Ohio, this sentence will have both knowledge \& Common Sense and temporal reasoning. This is because one should be able to tell that the birth location is in the northern hemisphere (knowledge) and December is part of the Winter in the northern hemisphere (temporal reasoning)).
\end{itemize}

\paragraph{Coreference}
This category includes cases where expressions refer to the same entity. However, we do not include the standard gamut of coreference phenomena in this category because the premise is not textual. We specifically include the following phenomena in this category:
Pronoun coreference, where the pronoun in a hypothesis refers to a noun phrase either in the hypothesis or the table. E.g., \example{Chris Jericho lives in a different state than he was born in.}
A noun phrase (not a named entity) in the hypothesis refers to a name of an entity in the table. For example, the table may say that \example{Bob has three children, including John} and the hypothesis says that \example{Bob has a son}. Here the phrase \example{a son} refers to the name \example{John}.

If there is a pronoun involved, we should not treat it as entity type or knowledge even though knowledge may be needed to know that, say, \example{Theresa May} is a woman and so we should use the pronoun \example{she}.

To avoid annotator confusion, when two names refer to each other, we label it only as the Named Entities category. For example, if the table talks  about \example{William Henry Gates III} and the hypothesis describes \example{Bill Gates}, even though the two phrases do refer to each other, we will label this as Named Entities. 

\paragraph{Quantification}
Any reasoning that involves introducing a quantifier such as every, most, many, some, none, at least, at most, etc. in the hypothesis. This category also includes cases where prefixes such as multi- (e.g., \example{multi-ethnic}) are used to summarize multiple elements in the table.

To avoid annotator confusion, we decide that the mere use of quantifiers like most and many is quantification. However, if the quantifier is added after comparing two numerical values in the table, the sentence is labeled to have numerical reasoning as well.

\paragraph{Subjective/Out of table}
Subjective inferences refer to any inferences that involve either value judgment about a proposition or a qualitative analysis of a numerical quantity. Out of table inferences involve hypotheses that use extra knowledge that is neither a well known universal fact nor common sense.  Such hypotheses may be written as factive or implicative constructions. Below are some examples of this category:

\begin{itemize}
    \item Based on a table about Chennai: \example{Chennai is a very good city.}
    \item If the table says that John's height is 6 feet, then the hypothesis that \example{John is a tall person.} may be subjective. However, if John's height is 8 feet tall, then the statement that \example{John is tall.} is no longer subjective, but common sense.
    \item If the table only says that John lived in Madrid and Brussels, and the hypothesis is \example{John lived longer in Madrid than Brussels.} This inference involves information that is neither well known nor common sense.

\item Based on the table of the movie Jaws, the hypothesis \example{It is known that Spielberg directed Jaws} falls in this category. The table may contain the information that Spielberg was the director, but this may or may not be well known. The latter information is out of the table. 
\end{itemize}

\paragraph{Syntactic Alternations}
This refers to a catch-all category of syntactic changes to phrases. This includes changing the preposition in a PP, active-passive alternations, dative alternations, etc. We expect that this category is rare because the premise is not text. However, since there are some textual elements in the tables, the hypothesis could paraphrase them.

This category is different from reasoning about named entities. If a syntactic alternation is applied to a named entity (e.g., \example{The Baltimore City Police} being written as \example{The Police of Baltimore City}), we will label it as a Named Entity if, and only if, we consider both phrases as named entities. Otherwise, it is just a syntactic alternation. Below are some examples of this category:

\begin{itemize}
    \item \example{New Orleans police officer} being written as \example{police officer of New Orleans}.
    
    \item \example{Shakespeare's sonnet} being written as \example{sonnet of Shakespeare}.
\end{itemize}

\paragraph{Ellipsis}
This category is similar in spirit to the category Ellipsis/Implicits in GLUE: ``An argument of a verb or another predicate is elided in the text, with the reader filling in the gap.'' Since in our case, the only well-formed text is in the hypothesis, we expect such gaps only in the hypothesis. (Compared to GLUE, where the description makes it clear that the gaps are in the premises and the hypotheses are constructed by filling in the gaps with either correct or incorrect referents.). For example, in a table about Norway that lists the per capita income as $\$$74K, the hypothesis that \example{The per capita income is $\$$74K.} elides the fact that this is about citizens of Norway, and not in general. %


\section{\datasetName Worker Analysis}
\label{sec:appendix_worker_stat}

\begin{figure}[!ht]
    \centering
    \includegraphics[width=0.48\textwidth]{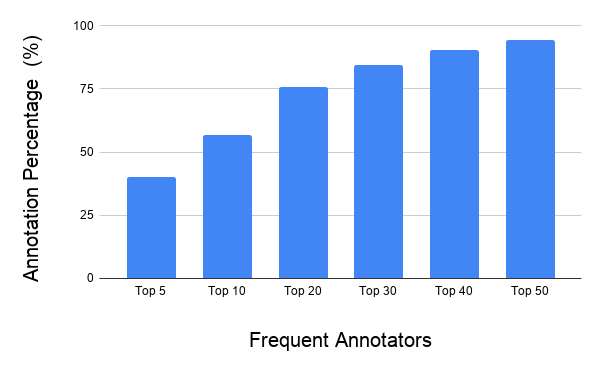}
    \caption{Number of annotations by frequent annotators}
    \label{fig:anno_stats_gen}
\end{figure}

Figure \ref{fig:anno_stats_gen} shows the number of examples annotated by frequent top-$n$ workers. We can see that the top 40 annotators annotated about 90$\%$ of the data. This observation is concordant with other crowd-sourced data annotation projects such as SNLI and MultiNLI~\cite{gururangan2018annotation}.

\section{\datasetName Dataset Statistics}
\label{sec:generation_stats_appendix}
In this section, we provide some essential statistics that will help in a better understanding of the dataset.

Table \ref{tab:stats_tables} shows a split-wise analysis of premises and annotators. The table shows that there is a huge overlap between the train set and the other splits except \alphaThree. This is expected since \alphaThree~ is from a different domain. Also, we observe that tables in \alphaThree~ are longer. In the case of annotators, we see that most of our dataset across all splits was annotated by the same set of annotators. 

\begin{table*}[!ht]
    \centering
    \begin{tabular}{lrrrrrr}
        \toprule
        Split & Train & Dev &\bf \alphaOne &\bf \alphaTwo &\bf \alphaThree  \\ \midrule
        Number of Unique Keys & 1558 & 411 & 466 & 332 & 409 \\ 
        Number of Unique Keys Intersection with Train & -  &334 & 312 & 273 & 94\\ 
        Average $\#$ of keys per table & 8.8 & 8.7 & 8.8 & 8.8 & 13.1\\
        
        Number of Distinct Annotators & 121 & 35 & 37 & 31 & 23 \\ 
        Annotator Intersection with Train & - & 33 & 37 & 30 & 19 \\ 
        Number of  Instances annotated by a Train annotator & - & 1794 & 1800 & 1797 & 1647 \\ 
        \bottomrule
    \end{tabular}
    \caption{Statistics of the premises and annotators across all discussed train-test splits}
    \label{tab:stats_tables}
\end{table*}

Table \ref{tab:hyp_stats} presents information on the generated hypotheses. The table lists the average number of words in the hypotheses. This is important because a dissimilar mean value of words would induce the possibility of length bias, i.e., the length of the sentences would be a strong indicator for classification. 

\begin{table}[!ht]
    \centering
    \begin{tabular}{lrrrrr}
    \toprule
    Label  & Train & Dev &\alphaOne &\alphaTwo &\alphaThree \\ \midrule
    Entail & 9.80 & 9.71 & 9.90 & 9.33 & 10.5 \\ 
     Neutral & 9.84 & 9.89 & 10.05 & 9.59 & 9.84 \\ 
    Contradict & 9.37 & 9.72 & 9.84 & 9.40 & 9.86 \\ 
    \bottomrule
    \end{tabular}
\caption{Mean length of the generated hypothesis sentences across all discussed train-test splits (standard deviation is in range 2.8 to 3.5)}
\label{tab:hyp_stats}
\end{table}

Table \ref{tab:hyp_prem_overlap_stats} shows the overlap between hypotheses and premise tables across various splits. Stop words like \emph{a, the, it, of,} etc. are removed. We observe that the overlap is almost similar across labels.

\begin{table}[!ht]
    \centering
    \begin{tabular}{lrrrrrr}
    \toprule
    Label & Train & Dev &\bf \alphaOne &\bf \alphaTwo &\bf \alphaThree \\ \midrule
    Entail &  0.52 & 0.47 & 0.45 & 0.46 & 0.48\\
    Neutral & 0.46 & 0.44 & 0.44 & 0.49 & 0.46\\
    Contradict & 0.44 & 0.43 & 0.45 & 0.44 & 0.46\\ 
     \bottomrule
    \end{tabular}
\caption{Mean statistic of the hypothesis sentences word overlapped with premises tables across all discussed train-test splits (standard deviation is in range 0.17 to 0.22)}
\label{tab:hyp_prem_overlap_stats}
\end{table}

Table \ref{tab:main_categories} and \ref{tab:alphaThree_categories} show the distribution of table categories in each split. We accumulate all the categories occurring for less than 3\% for every split into the ``Other'' category.

\begin{table}[!ht]
    \centering
    \begin{tabular}{lrrrr}
    \toprule
    Category & Train & Dev &\bf \alphaOne &\bf \alphaTwo \\ \midrule
    Person & 23.68 & 27 & 28.5 & 35.5 \\ 
    Musician & 14.66 & 19 & 18.5 & 22.5 \\ 
    Movie & 10.17 & 10 & 9 & 11.5 \\ 
    Album & 9.08 & 7 & 3.5 & 4.5 \\ 
    City & 8.05 & 8.5 & 8 & 7 \\ 
    Painting & 5.98 & 4.5 & 4 & 3.5 \\ 
    Organization & 4.14 & 2 & 1 & 0.5 \\ 
    Food / Drinks & 4.08 & 4 & 4 & 3 \\ 
    Country & 3.74 & 6 & 9 & 3.5 \\ 
    Animal & 3.56 & 4.5 & 4 & 4 \\ 
    Sports & 4.6 & 3.5 & 2.5 & 0.0 \\ 
    Book & 2.18 & 0.5 & 3 & 2.5 \\ 
    Other & 6.07 & 8.00 & 5.00 & 2.00\\ \bottomrule
    \end{tabular}
    \caption{Categories for all data splits (excluding \alphaThree) in percentage ($\%$). Others ($< 3\%$) include categories such as University, Event, Aircraft, Product, Game, Architecture, Planet, Awards, Wineyard, Airport, Language, Element, Car}
    \label{tab:main_categories}
\end{table}

\begin{table}[!ht]
\centering
\begin{tabular}{lr}
    \toprule
    Category & \alphaThree~ ($\%$) \\ \midrule
    Diseases & 20.4\\ 
    Festival & 17.41\\ 
    Bus / Train Lines & 14.93\\ 
    Exams & 8.46\\ 
    Element & 4.98\\ 
    Air Crash & 3.98\\ 
    Bridge & 3.98\\ 
    Disasters & 3.48\\ 
    Smartphone & 3.48\\ 
    Other & 18.9\\ \bottomrule
\end{tabular}
\caption{Categories for \alphaThree~ datasplit. Others ($< 3\%$) include categories such as Computer, Occupation, Restaurant, Engines, Equilibrium, OS, Cloud, Bus/Train Station, Coffee House, Cars, Bus/Train Provider, Hotel, Math, Flight}
\label{tab:alphaThree_categories}
\end{table}


\section{F1 Score Analysis}
\label{sec:f1score}

The F1 scores per label for two model baselines are in Table \ref{tab:f1Score}. We observe that neutral is easier than entailment and contradiction for both baseline, which is expected as neutrals are mostly associated with subjective/out-of-table reasonings which makes them syntactically different and easier to predict correctly. Despite this, we found that in all evaluations in (\S\ref{sec:experiments}) (except for \alphaTwo~test set), our models found neutrals almost as hard as the other two labels, with only an $\sim 3\%$ gap between the F-scores of the neutral label and the next best label. For \alphaTwo~test set neutral are much easier than entailment and contradiction. This is expected as entailment and contradiction in \alphaTwo~were adversarially flipped; hence, these predictions become remarkably harder compared to neutrals. Furthermore, \alphaThree~is the hardest data split, followed by \alphaTwo~and \alphaOne.  

\begin{table}
    \centering
    \begin{tabular}{cccc}
    \toprule
    \multicolumn{4}{c}{Premise as Paragraph}  \\\midrule
    Split & Entailment & Neutral & Contradiction \\ \midrule
    Dev & 76.19 &  79.02 & 72.73 \\
    \alphaOne & 74.69 & 77.85 & 69.85 \\
    \alphaTwo & 57.06 & 80.36 & 62.14 \\
    \alphaThree & 65.27 & 66.06 & 61.61 \\ \midrule
    \multicolumn{4}{c}{ Premise as TabFact} \\ \midrule
    Split & Entailment & Neutral & Contradiction \\ \midrule
    Dev & 77.69 & 79.45 & 74.77 \\
    \alphaOne & 76.43 & 80.34 & 73.07 \\
    \alphaTwo & 55.34 & 80.83 & 64.44 \\ 
    \alphaThree & 65.92 & 67.28 & 63.57 \\ \bottomrule
    \end{tabular}
    \caption{F1 Score ($\%$) with various baselines. All models are trained with RoBERTa$_{L}$ }
    \label{tab:f1Score}
\end{table}

\section{Statistics of \datasetName Verification}
\label{sec:appendix_verification}

Table \ref{tab:exact_verification_stats} shows the detailed agreement statistics of verification for the development and the three test splits. For every premise-hypothesis pair, we asked five annotators to verify the label. The table details the verification agreement among the annotators, and also reports how many of these majority labels match the gold label (i.e., the label intended by the author of the hypothesis). We also report individual annotator label agreement by matching the annotator's label with the gold label and majority label for an example. Finally, the table reports the Fleiss Kappa (across all five annotation labels) and the Cohen Kappa (between majority and gold label) for the development and the three test splits.

We see that, on average, about 84.8$\%$ of individual labels match with the majority label across all verified splits. Also, an average of 75.15$\%$ individual annotations also match the gold label across all verified splits.

From Table \ref{tab:exact_verification_stats}, we can calculate the percentage of examples with at least 3, 4, and 5 label agreements across 5 verifiers for all splits. For all splits, we have very high inter-annotator agreement of $>$95.85$\%$ for at-least 3, $>$ 74.50$\%$ for at-least 4 and 43.91$\%$ for at-least 5 annotators. The number of these agreements match with the gold label are: $>$81.76$\%$ for at-least 3, $>$ 67.09$\%$ for at-least 4 and 40.85$\%$ for at-least 5 for all splits.

\begin{table}[!ht]
    \centering
    \begin{tabular}{lcc}
    \hline \toprule
    \multicolumn{3}{c}{Exact agreement between annotators} \\ \midrule
    \bf Dataset &\bf  Number  &\bf  Gold/Total \\\midrule
    & 3 & 350 / 469\\
    Dev & 4 & 529 / 601 \\
    & 5 & 550 / 605 \\
    & no agreement & 116 \\ \midrule
    & 3 & 184 / 292\\
    \alphaOne & 4 & 459 / 533 \\
    & 5 & 863 / 922 \\
    & no agreement & 45 \\ \midrule 
        & 3 & 245 / 348\\
    \alphaTwo & 4 & 453 / 537 \\
    & 5 & 812 / 857 \\
    & no agreement & 58 \\ \midrule
            & 3 & 273 / 422\\
    \alphaTwo & 4 & 441 / 524 \\
    & 5 & 706 / 765 \\
    & no agreement & 79 \\ \bottomrule \toprule
    \multicolumn{3}{c}{Individual agreement with gold / majority label} \\\midrule
    \bf Dataset &\bf  Statistics  &\bf  Agreement ($\%$) \\\midrule
    Dev & Gold & 71.12 \\
    & Majority & 81.65 \\ \midrule 
        \alphaOne & Gold & 78.52 \\
    & Majority & 87.24 \\ \midrule 
        \alphaTwo & Gold & 77.74 \\
    & Majority & 86.32 \\ \midrule 
            \alphaThree & Gold & 73.22 \\
    & Majority & 84.01 \\ \midrule
        Average & Gold & 75.15\\
    & Majority & 84.8\\\bottomrule \toprule
    \multicolumn{3}{c}{Kappa values across splits} \\ \midrule
    \bf Dataset &\bf  Fleiss  &\bf  Cohen \\\midrule
    Dev & 0.4601  & 0.7793 \\ 
    \alphaOne &  0.6375 & 0.7930 \\ 
    \alphaTwo & 0.5962 & 0.8001 \\
    \alphaThree & 0.5421 & 0.7444 \\ \midrule \bottomrule
    \end{tabular}
    \caption{Exact, Individual and Kappa values for verification's statistics.}
    \label{tab:exact_verification_stats}
\end{table}




\end{document}